\documentclass[10pt,twocolumn,letterpaper]{article}

\usepackage[pagenumbers]{cvpr} 

\usepackage{gensymb}
\usepackage{soul}

\definecolor{cvprblue}{rgb}{0.21,0.49,0.74}
\usepackage[pagebackref,breaklinks,colorlinks,allcolors=cvprblue]{hyperref}

\usepackage{times}
\usepackage{url}
\usepackage{soul}
\usepackage{amsmath}
\usepackage{multirow}
\usepackage{booktabs}
\usepackage{graphicx}
\usepackage{makecell}
\usepackage{subcaption}
\usepackage{pifont}
\usepackage{bm}
\usepackage{footmisc}
\newcommand{\sysname}{FLASH}

\def\paperID{***} 
\def\confName{CVPR}
\def\confYear{2025}

\title{Learning to Animate Images from A Few Videos to Portray Delicate Human Actions}

\author{Haoxin Li\textsuperscript{\rm 1}\thanks{Part of the work is done during an internship at ByteDance.},
Yingchen Yu\textsuperscript{\rm 2},
Qilong Wu\textsuperscript{\rm 3},
Hanwang Zhang\textsuperscript{\rm 1},
Song Bai\textsuperscript{\rm 2},
Boyang Li\textsuperscript{\rm 1}
\\
\\
\textsuperscript{\rm 1}{Nanyang Technological University}\quad \textsuperscript{\rm 2}{ByteDance}\quad \textsuperscript{\rm 3}{National University of Singapore}
}

\begin{document}
\maketitle
\begin{abstract}
Despite recent progress, video generative models still struggle to animate static images into videos that portray delicate human actions, particularly when handling uncommon or novel actions whose training data are limited.
In this paper, we explore the task of learning to animate images to portray delicate human actions using a small number of videos---16 or fewer---which is highly valuable for real-world applications like video and movie production.
Learning generalizable motion patterns that smoothly transition from user-provided reference images in a few-shot setting is highly challenging.
We propose \sysname{} (\textbf{F}ew-shot \textbf{L}earning to \textbf{A}nimate and \textbf{S}teer \textbf{H}umans), which learns generalizable motion patterns by forcing the model to reconstruct a video using the motion features and cross-frame correspondences of another video with the same motion but different appearance. This encourages transferable motion learning and mitigates overfitting to limited training data. Additionally, \sysname{} extends the decoder with additional layers to propagate details from the reference image to generated frames, improving transition smoothness. 
Human judges overwhelmingly favor \sysname{}, with 65.78\% of 488 responses prefer \sysname{} over baselines. We strongly recommend watching the videos in the Webpage\footnote{\scriptsize{\url{https://lihaoxin05.github.io/human_action_animation/}}}, as motion artifacts are hard to notice from images.
\end{abstract}    
\section{Introduction}
\label{sec:introduction}
\begin{figure*}
\setlength{\abovecaptionskip}{0.5em}
\centering
\includegraphics[width=\linewidth]{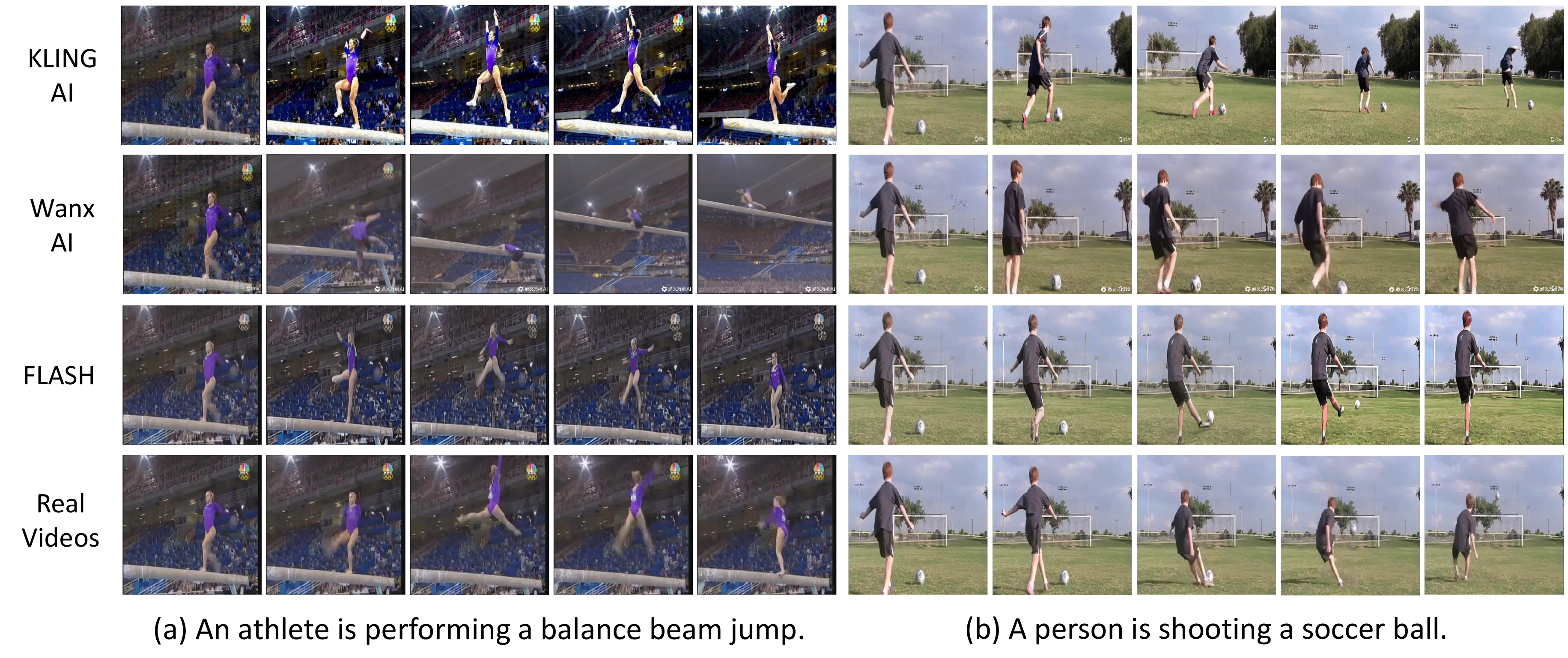}
\caption{Comparison of animated human action videos produced by KLING AI, Wanx AI and \sysname{} (our method). In the balance beam jump action, Wanx AI produces physics-defying movements, whereas KLING AI generates a jump but fails to portray the standard jump on the balance beam. For the soccer shooting action, both KLING AI and Wanx AI struggle to generate the correct shooting motion and the person never kicks the ball away. In contrast, \sysname{} successfully generates actions that resemble the real-world actions in the last row. We strongly recommend watching the animated videos in the Webpage, as motion artifacts can be hard to notice from static images.}
\label{fig:comparing-tools-1}
\vspace{-1em}
\end{figure*}

Despite substantial progress \citep{ho2022video,singer2022make,zhou2022magicvideo,guo2023animatediff,wang2023videofactory,esser2023structure,yin2023dragnuwa,liew2023magicedit,zhang2023magicavatar,he2023animate,wu2023tune,wang2024videocomposer,wang2024boximator,yang2024cogvideox,kong2024hunyuanvideo}, video generative models still struggle to accurately portray delicate human actions, especially when they are required to start from a user-provided reference image. Even commercial AI video generators trained on large-scale datasets, such as KLING AI\footnote{\scriptsize{\url{https://www.klingai.com/image-to-video}}} and Wanx AI\footnote{\scriptsize{\url{https://tongyi.aliyun.com/wanxiang/videoCreation}}}, encounter difficulty with this task. As shown in Figure \ref{fig:comparing-tools-1}, both fail to animate actions such as balance beam jump or shooting a soccer ball. 

We attribute the difficulty mainly to the interplay between action complexity and training data scarcity. Due to skeletal and joint structures, human actions are highly complex and constrained in unique ways. For example, the range of forearm pronation is typically 80-90\degree, whereas the neck can do 150\degree{} rotation and 125\degree{} flexion \cite{Chen1999:Spine}. Inferring such precise specifications from videos alone is a difficult inverse problem. To make things worse, human action videos follow a long-tailed distribution \cite{zhang2021videolt}, which means that, for a wide range of human actions, only a small amount of data are available for learning. The action recognition community has long recognized this problem and created a body of literature on few-shot and open-vocabulary action recognition \cite{zhang2020few,perrett2021temporal,huang2024froster,lin2024rethinking}. Interestingly, few-shot generation of human actions remains under-investigated. 




In this paper, we explore the task of animating an image to portray delicate human actions by learning from a small set of videos. The problem input includes a static reference image as well as a textual prompt describing the action. We learn from up to 16 videos for each action class, thereby reducing the need for extensive video data collection. The animation should begin exactly with the reference image, giving users precise control over the initial state of the video. This capability  is particularly valuable for applications like video and movie production, which need to animate diverse actors performing uncommon or newly designed actions, often with only a few example videos available and requiring precise control over the initial states of the action, such as the actors' orientations and spatial arrangement of the scenes. Controllable generation has been studied for both images  \cite{pan2023drag,shi2023dragdiffusion,zhang2023adding} and videos \cite{Hao_2018_CVPR,yin2023dragnuwa,ma2024trailblazer}, but the additional constraint also renders the problem more difficult. 

Existing image animation methods encounter considerable difficulties with this task. These approaches typically rely on large video datasets for training and primarily focus on preserving the appearance of the reference images \citep{xing2023dynamicrafter,guo2023i2v,jiang2023videobooth,gong2024atomovideo,wang2023dreamvideo,guo2023sparsectrl,ma2024follow,ren2024consisti2v,gong2024atomovideo,zhang2023pia,guo2024i2v} or learning spatial-temporal conditioning controls (\eg, optical flows) to guide image animation \citep{ni2023conditional,kandala2024pix2gif,shi2024motion}. However, with limited training data, these methods suffer from overfitting and struggle to learn generalizable motion patterns. Customized video generation methods \citep{materzynska2023customizing,wei2024dreamvideo,zhao2023motiondirector,li2024motrans} can learn target motion from a few examples but fall short in the ability to align the action  with the user-provided reference image.

We propose \sysname{} (\textbf{F}ew-shot \textbf{L}earning to \textbf{A}nimate and \textbf{S}teer \textbf{H}umans), a framework for few-shot human action animation. To learn generalizable motion patterns from limited videos, \sysname{} introduces the Motion Alignment Module, which forces the model to reconstruct a video using the motion features of another video with the same motion but different appearance. This facilitate the learning of generalizable motion features and reduces overfitting. Additionally, \sysname{} employs a Detail Enhancement Decoder to propagate multi-scale details from the reference image to the generated frames, providing smooth transition from the first reference frame.

Experiments on 16 delicate human actions demonstrate that \sysname{} accurately and plausibly animates diverse reference images into lifelike human action videos. Human judges overwhelmingly favor \sysname{}; 65.78\% of 488 responses prefer \sysname{} over open-source baselines. 
\sysname{} also outperforms existing methods across automatic metrics, and generalizes to non-realistic figures like cartoon characters or humanoid aliens.


\section{Related Work}
\label{sec:related_work}
\noindent\textbf{Video Generation.}
Video generation using diffusion models \citep{ho2020denoising,song2020score,song2020denoising} have notably surpassed methods based on GANs \citep{goodfellow2020generative}, VAEs \citep{kingma2013auto} and flow techniques \citep{chen2019residual}. Diffusion models for video generation can be broadly classified into two groups. The first group generates videos purely from text descriptions. These methods extend advanced text-to-image generative models by integrating 3D convolutions, temporal attention layers, or 3D full attention layers to capture temporal dynamics in videos \citep{ho2022video,ho2022imagen,singer2022make,zhou2022magicvideo,blattmann2023align,guo2023animatediff,wang2023videofactory,yang2024cogvideox,kong2024hunyuanvideo}. To mitigate concept forgetting when training on videos, some methods use both videos and images jointly for training \citep{ho2022video,chen2024videocrafter2,kong2024hunyuanvideo}. Large Language Models (LLMs) contribute by generating frame descriptions \citep{gu2023seer,huang2024free,li2024vstar} and scene graphs \citep{fei2023empowering} to guide the video generation. Trained on large-scale video-text datasets \citep{bain2021frozen,xue2022advancing,chen2024panda}, these methods excel at producing high-fidelity videos. However, they typically lack control over frame layouts like object positions. To improve controllability, LLMs are used to predict control signals \citep{lu2023flowzero,lian2023llm,lv2024gpt4motion}, but these signals typically offer coarse control (\eg, bounding boxes) rather than fine-grained control (\eg, human motion or object deformation).

On top of text descriptions, the second group of techniques uses additional control sequences, such as depth maps, optical flows, trajectories and bounding boxes \citep{esser2023structure,yin2023dragnuwa,liew2023magicedit,zhang2023magicavatar,he2023animate,zhang2024tora,wang2024videocomposer,wang2024boximator,wang2024motionctrl,hu2024animate,ma2024trailblazer}, to control frame layouts and motion. Additionally, several techniques use existing videos as guidance to generate videos with different appearances but identical motion \citep{wu2023tune,qi2023fatezero,yang2023rerender,geyer2023tokenflow,yang2024eva,zhang2023motioncrafter,ling2024motionclone,ren2024customize,park2024spectral,jeong2024vmc,xiao2024video}. However, these methods cannot create novel videos that share the same motion class with the guidance video but differ in the actual motion, such as human positions and viewing angles, which limits their generative flexibility.

\vspace{0.1em}\noindent\textbf{Image Animation.} Image animation involves generating videos that begin with a given reference image controlling the initial action states. Common approaches achieve this by integrating the image features into videos through cross-attention layers \citep{wang2023dreamvideo,xing2023dynamicrafter,jiang2023videobooth,gong2024atomovideo}, employing additional image encoders \citep{guo2023sparsectrl,guo2023i2v,wang2024microcinema}, or incorporating the reference image into noised videos \citep{zeng2023make,wu2023lamp,girdhar2023emu,ma2024follow,ren2024consisti2v,gong2024atomovideo}. Another line of methods focuses on learning guidance sequences (\eg, motion maps) that aligns with the reference image to guide the generation of subsequent frames \citep{shi2024motion,ni2023conditional,kandala2024pix2gif}. However, these approaches often require extensive training videos to learn motion or guidance sequences, making them ineffective with limited data.

\vspace{0.1em}\noindent\textbf{Customized Generation.} Customized generation creates visual content tailored to specific concepts using limited samples. In the image domain, static concepts are associated with new texts \citep{gal2022image,ruiz2023dreambooth,shi2024instantbooth,kumari2023multi} or model parameters \citep{chen2023subject,smith2023continual,gu2023mix}. In the video domain, \cite{molad2023dreamix,materzynska2023customizing,wei2024dreamvideo,zhao2023motiondirector,li2024motrans} learn target appearance and motion from limited data but lack control over initial action states, making it difficult for users to control the positions and directions of the actor and objects. Additionally, their requirement for test-time training on each reference image limits flexibility. While \cite{wu2023lamp,kansy2024reenact} are similar to our work in learning specific motion patterns from a few videos, they rely on the model to automatically prioritize motion over appearance, which limit the generalizability due to the lack of explicit guidance for appearance-general motion. In contrast, our work learns generalizable motion from a few videos with explicit guidance, enabling it to generalize to reference images with varying visual attributes, such as actor positions and textures.

\begin{figure*}[!t]
\setlength{\abovecaptionskip}{0.5em}
\centering
\includegraphics[width=\linewidth]{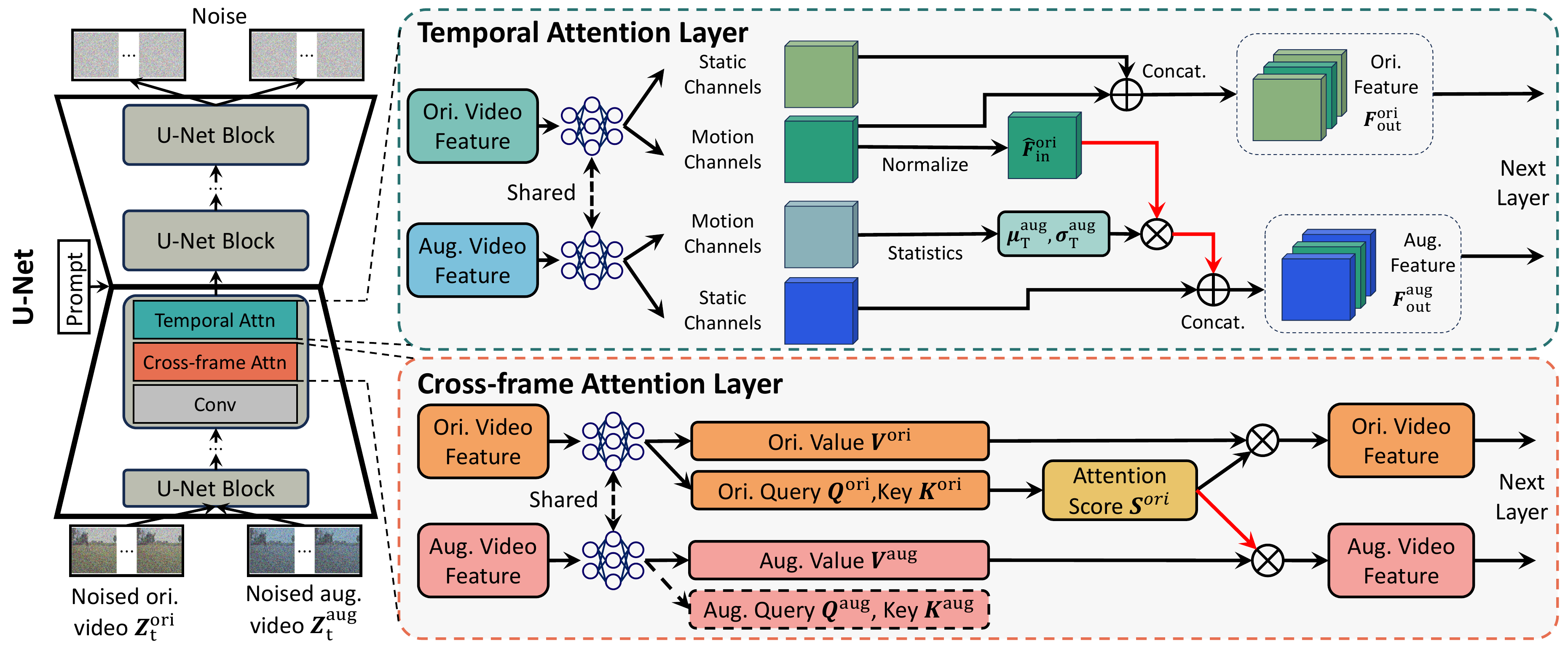}
\caption{An illustration of the Motion Alignment Module. Both the noised latent representations of the original and strongly augmented videos are input to the U-Net. In the temporal attention layers, static and motion features are extracted from both videos. Motion features from the original video are transferred to the augmented video (red arrows), and the recombined features are passed to the next layer. In the cross-frame attention layers, attention scores from the original video, which capture its cross-frame motion structure, are used to warp the augmented video (red arrow) before passing it to the next layer. The U-Net is trained to predict the noise added to both videos based on the motion patterns of the original video, encouraging the learning of consistent motion patterns.}
\label{fig:framework}
\vspace{-1em}
\end{figure*}

\section{\sysname{}}
\label{sec:method}
Building upon latent video diffusion (Sec. \ref{subsec:pre}), we propose two novel system components. The first is the Motion Alignment Module, detailed in Sec. \ref{subsec:motion_align}, which encourages the learning of generalizable motion patterns and prevents overfitting to static features. The second is the Detail Enhancement Decoder, explained in Sec. \ref{subsec:detail-decoder}, which propagates details from the user-provided reference image to generated frames to enhance transition smoothness.

\subsection{Preliminaries}
\label{subsec:pre}
\textbf{Latent Image Diffusion Model.} Latent Diffusion Model (LDM) \citep{rombach2022high} comprises four main components: an image encoder $\mathcal{E}$, an image decoder $\mathcal{D}$, a text encoder $\mathcal{T}$, and a U-Net $\epsilon_\theta$. The training begins with an image $\boldsymbol{x}$ and a textual description $y$. We encode the image into a latent representation $\boldsymbol{z}_0=\mathcal{E}(\boldsymbol{x})\in\mathbb{R}^{h \times w \times c}$. Next, for a randomly sampled time index $t$, we add Gaussian noise $\boldsymbol{\epsilon}_t\sim\mathcal{N}(\boldsymbol{0},\,I)$ to the latent, yielding a noised version $\boldsymbol{z}_t=\sqrt{\Bar{\alpha}_t}\boldsymbol{z}_0+\sqrt{1-\Bar{\alpha}_t}\boldsymbol{\epsilon}_t$, where $\Bar{\alpha}_t$ is the noise strength \citep{dhariwal2021diffusion,ho2020denoising}. The main step is to train the U-Net $\epsilon_\theta(\boldsymbol{z}_t, t, \mathcal{T}(y))$ to predict the added noise $\bm{\epsilon}_t$, so that it can be subtracted to recover the original latent $\boldsymbol{z}_0$. Since identifying the noise and identifying the original latent representation are two sides of the same coin, we often refer to the task of the U-Net as reconstructing the original latent representation. 
During inference, we randomly sample a latent noise $\boldsymbol{z}_T \sim \mathcal{N}(\boldsymbol{0},\,I)$ and progressively perform denoising to obtain an estimated noise-free latent image $\hat{\boldsymbol{z}}_0$ using the trained U-Net. Finally, the decoder recovers the generated image in pixel space $\hat{\boldsymbol{x}}=\mathcal{D}(\hat{\boldsymbol{z}}_0)$.


\vspace{0.2em}\noindent\textbf{Latent Video Diffusion Model.} The LDM framework can be naturally extended to video generation. Given a video consisting of $N$ frames $\boldsymbol{X}=\langle\boldsymbol{x}^i\rangle_{i=1}^{N}$, we encode each frame and yield a latent video representation $\boldsymbol{Z}_0=\langle\boldsymbol{z}_0^i\rangle_{i=1}^{N} \in \mathbb{R}^{N \times h \times w \times c}$. The training is the same as before with the following loss:
%
\begin{equation}\label{eq:video-ldm}
    \mathcal{L}_D=\mathbb{E}_{\boldsymbol{X}, y, \boldsymbol{\epsilon}_t \sim \mathcal{N}(\boldsymbol{0},\,I), t}
    \left[\left\|\boldsymbol{\epsilon}_t-\epsilon_\theta\left(\boldsymbol{Z}_t, t, \mathcal{T}(y)\right)\right\|^2_2\right].
\end{equation} 
The video denoising U-Net differs from the image counterpart in a few ways. 
To capture temporal dynamics, \citep{ho2022video,esser2023structure,guo2023animatediff,guo2023sparsectrl} add temporal attention after each spatial attention layer. To enhance consistency with the reference frame, \citep{text2video-zero,wu2023lamp} replace spatial self-attention in the U-Net with spatial cross-frame attention, where features from the reference frame (typically the first frame) serve as the keys and values. These layers propagate appearance features from the reference frame to other frames to improve consistency. Further, the reference frame is kept noise-free in the noised latent video to preserve its appearance \citep{wu2023lamp,ren2024consisti2v}. \sysname{} incorporates all these components. Further details are in Appendix \ref{appendix:pre}.

\subsection{Motion Alignment Module}
\label{subsec:motion_align}
The Motion Alignment Module forces the model to learn motion patterns that remain consistent despite appearance changes. This is achieved by four steps. First, given a training video, we create a strongly augmented video with the same motion but different appearances. Second, we identify feature channels that capture motion information and static information from each video. We recombine the motion features of the original video and static features of the augmented video as the new features of the augmented video. Third, we force the cross-frame spatial attention in the augmented video to adhere to the attention in the original video, which aligns the motion structures of the two videos, as reflected by the attention weights. Finally, we train the network to reconstruct the latent (\ie predicting the added noise) of the augmented video using the the motion features and structures of the original video, which encourages the learned motion to be generalizable across different appearances.
The overall process is depicted in Figure \ref{fig:framework} and elaborated below.


\vspace{0.2em}\noindent\textbf{Strongly Augmented Videos.} Given an original video, $\boldsymbol{X}^{\text{ori}}$, we create a strongly augmented version $\boldsymbol{X}^{\text{aug}}$, which has different appearances but the same motion. We choose the augmentations as Gaussian blur with random kernel sizes and random color adjustments. The randomness of augmentation ensures that the model encounters different original-augmented video pairs at different training epochs. Details of the augmentations and example augmented videos are in Appendix \ref{appendix:strong_aug_video}. 



\vspace{0.2em}\noindent\textbf{Identifying Motion Channels.} 
We set out to identify motion features in the U-Net from the original video. We denote the features extracted by a temporal attention layer as $\boldsymbol{F}_{\text{in}}\in\mathbb{R}^{N \times h' \times w' \times c'}$, and compute its mean $\bm{\mu}_{\text{T}} \in \mathbb{R}^{h' \times w' \times c'}$ and standard deviation $\bm{\sigma}_{\text{T}} \in \mathbb{R}^{h' \times w' \times c'}$ along the temporal dimension.
\citep{xiao2024video} shows that motion information is predominantly encoded in a few channels. Thus, we take the average of $\bm{\sigma}_{\text{T}}$ across spatial positions, denoted as $\boldsymbol{s}\in\mathbb{R}^{c'}$, and consider channels in the top $\tau$-percentile as motion channels. We identify the channel indices from the original video and consider the corresponding channels in the augmented video also as motion channels. 

\vspace{0.2em}\noindent\textbf{Transferring Motion Channels.} The next step is to transfer motion channels from the original video to the augmented video, in order to encourage the network learn the motion channels generalizable to both videos. Before transferring, we remove the static components from $\boldsymbol{F}_{\text{in}}$ by normalization:
\begin{equation}
    \hat{\boldsymbol{F}}_{\text{in}}=\frac{\boldsymbol{F}_{\text{in}}-\bm{\mu}_{\text{T}}(\boldsymbol{F}_{\text{in}})}{\bm{\sigma}_{\text{T}}(\boldsymbol{F}_{\text{in}})}.
\end{equation}
We posit that normalization by the standard deviation reduces the influence of feature scales (\eg, varying brightness) but preserves motion. 


After that, with the motion channels identified from the original video, we replace the corresponding channels in the augmented video with those from the original, yielding a new feature map $\hat{\boldsymbol{F}}^{\text{aug}}_{\text{out}}$. Finally, we restore the video mean and standard deviation of the augmented video, $\boldsymbol{F}_{\text{out}}^{\text{aug}}=\hat{\boldsymbol{F}}_{\text{out}}^{\text{aug}} \bm{\sigma}_{\text{T}}^{\text{aug}}+\bm{\mu}_{\text{T}}^{\text{aug}}$, which are fed to the next layer and are eventually used in noise prediction. The original video features remain unchanged and are unaffected by this step.



\vspace{0.2em}\noindent\textbf{Cross-frame Attention Alignment.} The purpose of this technique is to guide the model to learn the same cross-frame motion structures from the two videos. Recall that the cross-frame attention treats the reference frame (the first frame) as the keys and values, and the current frame as the queries. Hence, the attention weights indicate how patches in the reference frame correspond to patches in the current frame. We want these correspondences to be identical in both videos. 


We denote the input features of a cross-frame attention layer as $\boldsymbol{F}_{\text{in}}=\langle\boldsymbol{f}^i_{\text{in}}\rangle_{i=1}^{N}\in\mathbb{R}^{N \times h' \times w' \times c'}$. The output features are computed as:
%
\begin{align}\label{eq:cross_frame_attention}
    \boldsymbol{F}_{\text{out}}&=\text{Softmax}\left(\frac{(\boldsymbol{Q}\boldsymbol{W}^Q)(\boldsymbol{K}\boldsymbol{W}^K)^\top}{\sqrt{c^{'}}}\right) (\boldsymbol{V}\boldsymbol{W}^V) \\ &=\boldsymbol{S} (\boldsymbol{V}\boldsymbol{W}^V),
\end{align}
where $\boldsymbol{Q}=\boldsymbol{F}_{\text{in}}$, $\boldsymbol{K}=\boldsymbol{f}^1_{\text{in}}$, $\boldsymbol{V}=\boldsymbol{f}^1_{\text{in}}$ are the query, key, and value, respectively, and $\boldsymbol{W}^Q$, $\boldsymbol{W}^K$, $\boldsymbol{W}^V$ are learnable matrices. The key and value are from the first frame of the video. Therefore, $\boldsymbol{S}$ represents the similarity between the query and the key from the first frame, which implicitly warps the first frame into subsequent frames \citep{mallya2022implicit}. 
Our technique simply applies the $\boldsymbol{S}$ computed from the original video when processing the augmented video, whose output features are used for noise prediction. 



\subsection{Detail Enhancement Decoder}
\label{subsec:detail-decoder}
The previous section introduces the Motion Alignment Module, which improves the generalization of the learned motion features in the latent space. However, a strong latent representation \emph{per se} does not guarantee the quality of the generated video; we still need a powerful decoder that can (1) reproduce natural motion in the decoded video, even though the latent representation may contain distortion or noise, and (2) ensure the later frames in the video flow from the first reference frame smoothly. 

To this end, we propose the Detail Enhancement Decoder. 
We introduce architectural components that directly retrieve multi-scale details from the reference frame and propagate them to later frames. Further, we apply strong data augmentation to the input, so that the decoder must learn to recover from distorted latent representations. 


\vspace{0.2em}\noindent\textbf{Multi-scale Detail Propagation.} We number the network layers in both the encoder $\mathcal{E}$ and decoder $\mathcal{D}$ using $l\in\{0,1,\cdots,L\}$. The features in the encoder are denoted as $\bm{g}_l$ and those in the decoder are denoted as $\bm{h}_l$. A special case is $\bm{g}_0$ and $\bm{h}_0$, which respectively denote the original and reconstructed video in pixel space. $\bm{g}_L$ is the output from the encoder and $\bm{h}_L$ is the input to the decoder; both are in the latent space. We further use superscripts to denote the frame number. $\boldsymbol{g}_{l}^1$ denotes the encoded feature from the first, user-supplied reference frame, taken from encoder layer $l$. A main goal of the Detail Enhancement Decoder is to propagate this feature to decoder features of later frames. 


In addition to the layer-by-layer structure of a typical decoder, we propose two new architectural components, namely the \emph{Warping Branch} and the \emph{Patch Attention Branch}.
The aim of the Warping Branch is to retrieve relevant appearance information from the reference frame $\boldsymbol{g}_{l}^1$ for each spatial location $(p, q)$ in the $i$-th frame $\boldsymbol{h}_{l}^i$. However, due to motion, it is not immediately clear which spatial location in $\boldsymbol{g}_{l}^1$ is relevant to position $(p, q)$ of the $i$-th frame. We simply apply a neural network $\mathcal{N}$ to predict the displacement $(\Delta p, \Delta q)$. The network takes the two frames $\boldsymbol{h}_{l}^i$, $\boldsymbol{g}_{l}^1$ as input, and outputs two scalars for each spatial location,
\begin{equation}
(\Delta p, \Delta q) = \mathcal{N}(\boldsymbol{h}_{l}^i, \boldsymbol{g}_{l}^1) [p, q].
\end{equation}
Hence, the relevant position from the reference frame is $(p+\Delta p, q+\Delta q)$. As $\Delta p$ and $\Delta q$ may not be integers, we perform bilinear interpolation to find the exact feature at that location. We store the retrieved features of all spatial locations in a new tensor $\bm{s}^i_l$.


To complement the local retrieval of the Warping Branch, we introduce the Patch Attention Branch, which retrieves details from the entire reference feature map $\boldsymbol{g}_{l}^1$.  We divide both $\boldsymbol{h}_{l}^i$ and $\boldsymbol{g}_{l}^1$ into patches and apply a standard cross-attention layer $\mathcal{A}$, using $\boldsymbol{h}_{l}^i$ as the query and $\boldsymbol{g}_{l}^1$ as both the key and the value. The output features are denoted as  $\boldsymbol{t}_{l}^i$.

We fuse the two output features using weights $\boldsymbol{w}^i_l$ produced by a network $\mathcal{M}$:
%
\begin{align}
&\boldsymbol{w}^i_l=\mathcal{M}(\boldsymbol{h}_{l}^i, \boldsymbol{g}_{l}^1),\\
&\hat{\boldsymbol{h}}_l^i=\boldsymbol{h}_l^i+\boldsymbol{w}^i_l\odot(\boldsymbol{s}_{l}^i+\boldsymbol{t}_{l}^i),
\end{align}
where $\odot$ represents element-wise multiplication. We pass the fused features $\hat{\boldsymbol{h}}_l^i$ to the next layer in the decoder, which is layer $l-1$.

\vspace{0.2em}\noindent\textbf{Distorted Latent Representation as Input.} Given an original video $\boldsymbol{X}^{\text{ori}}$, we first distort it into $\boldsymbol{X}^{\text{dis}}$ by applying Gaussian blur with a random kernel size, random color adjustments in randomly selected regions, and random elastic transformations. We then encode $\boldsymbol{X}^{\text{dis}}$ into a latent video $\boldsymbol{Z}^{\text{dis}}_0$. This process introduces potential distortions in latent videos. Note that the degree of distortion is randomly sampled, so there are some inputs to the decoder that receive minimal or zero distortion. 

\vspace{0.2em}\noindent\textbf{Reconstruction Loss.} The entire decoder, including the Warping Branch and the Patch Attention Branch, is end-to-end trainable. Therefore, we simply train the decoder to reconstruct $\boldsymbol{Z}^{\text{dis}}_0$ back to $\boldsymbol{X}^{\text{ori}}$. We denote the decoded video from $\boldsymbol{Z}^{\text{dis}}_0$ as $\hat{\boldsymbol{X}}^{\text{ori}}=\mathcal{D}(\boldsymbol{Z}^{\text{dis}}_0)$, and use a reconstruction loss to train the newly added layers (\ie, $\mathcal{N}$, $\mathcal{A}$ and $\mathcal{M}$):
%
\begin{equation}
    \mathcal{L}_R=\|\hat{\boldsymbol{X}}^{\text{ori}}-\boldsymbol{X}^{\text{ori}}\|^2_2.
\end{equation}
More detail are provided in Appendix \ref{appendix:decoder}.

\section{Experiments}
\label{sec:experiments}
We conduct experiments on 16 actions selected from the HAA500 dataset \citep{chung2021haa500}, including single-person actions (\textsf{\small push-up}, \textsf{\small arm wave}, \textsf{\small shoot dance}, \textsf{\small running in place}, \textsf{\small sprint run}, and \textsf{\small backflip}), human-object interactions (\textsf{\small soccer shoot}, \textsf{\small drinking from a cup}, \textsf{\small balance beam jump}, \textsf{\small balance beam spin}, \textsf{\small canoeing sprint}, \textsf{\small chopping wood}, \textsf{\small ice bucket challenge}, and \textsf{\small basketball hook shot}), and human-human interactions (\textsf{\small hugging human}, \textsf{\small face slapping}). Most of these classes are challenging for general video generative models to animate. We train a separate model for each action. More details about data and implementation are in Appendices \ref{appendix:data} and \ref{appendix:imple_detail}.

\begin{figure}[tb]
\centering
\includegraphics[width=\linewidth]{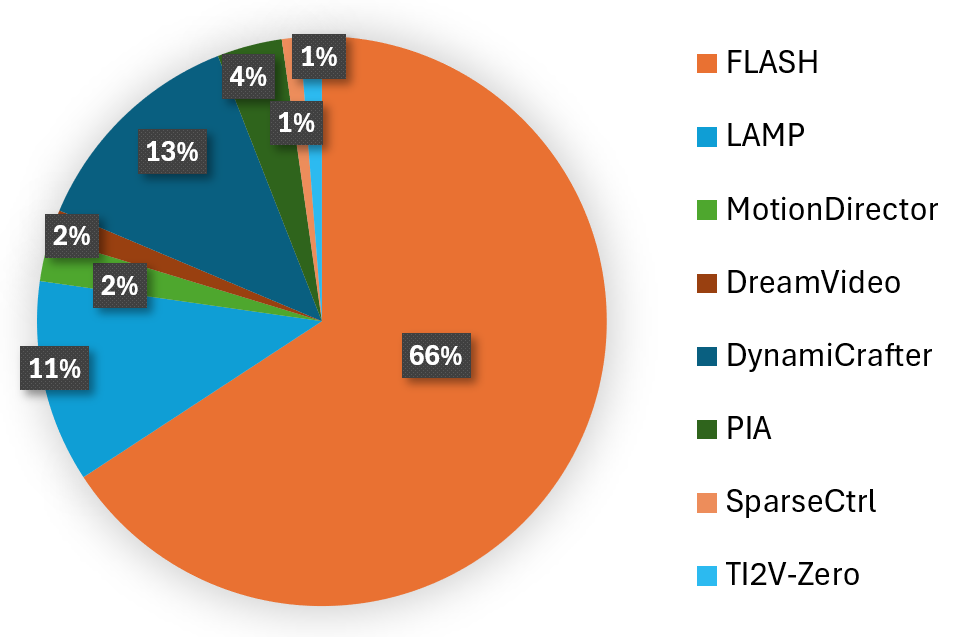}
\caption{The percentage of users that choose products of each video generator as the best videos in the user study on Amazon Mechanical Turk. The proposed method, \sysname{}, received the vast majority of votes. }
\label{fig:AMT_results}
\vspace{-0.5em}
\end{figure}

\begin{figure*}[tb]
\setlength{\abovecaptionskip}{0.5em}
\centering
\includegraphics[width=\linewidth]{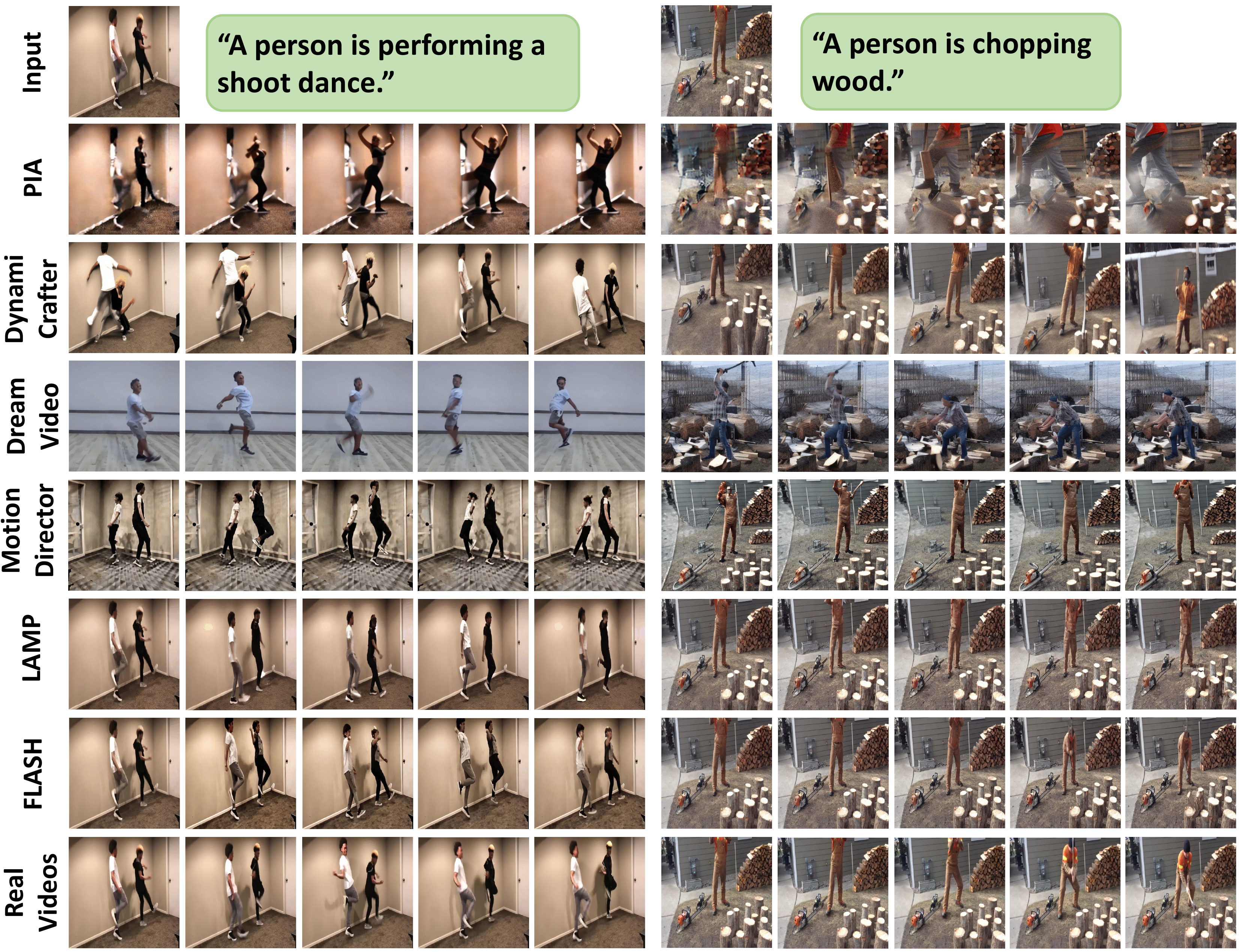}
\caption{Qualitative comparison of different methods.  We strongly recommend watching the animated videos in the Webpage, as motion artifacts are hard to notice from static images.}
\label{fig:results-cmp}
\vspace{-0.5em}
\end{figure*}

\subsection{Main Results}
We compare \sysname{} with several baselines, including TI2V-Zero \citep{ni2024ti2v}, SparseCtrl \citep{guo2023sparsectrl}, PIA \citep{zhang2023pia}, DynamiCrafter \citep{xing2023dynamicrafter}, DreamVideo \citep{wei2024dreamvideo}, MotionDirector \cite{zhao2024motiondirector} and LAMP \citep{wu2023lamp}, whose details are described in Appendix \ref{appendix:baselines}.

\vspace{0.1em}\noindent\textbf{User Study.} Existing automatic evaluation metrics fall significantly short of detecting all types of artifacts. Human evaluation remains the gold standard in evaluating the quality of generated videos. Thus, we perform a large-scale user study on Amazon Mechanical Turk to compared the videos generated by \sysname{} and baselines. 

In the user study, workers were tasked to select the video of the highest quality from a set of candidates. For each action, we randomly selected four different reference images and their corresponding generated videos for this user study. To identify random clicking, each question was paired with a control question that has obvious correct answer. The control question includes a real video of a randomly selected action alongside clearly incorrect ones, such as a static video. The main question and the control question were randomly shuffled within each question pair, and each pair was evaluated by 10 different workers. Responses from workers who failed the control questions were marked as invalid. More details are provided in Appendix \ref{appendix:user_study}. Among 488 valid responses, \sysname{} was preferred in 65.78\% of cases, as shown in Figure \ref{fig:AMT_results}, which significantly outperforms other methods and highlights its superiority.

\vspace{0.1em}\noindent\textbf{Qualitative Results.} In Figure \ref{fig:results-cmp}, we compare \sysname{} with PIA, DynamiCrafter, DreamVideo, MotionDirector and LAMP, while providing results for other methods on the Webpage due to space constraints. From the results, we observe that \emph{PIA} and \emph{DynamiCrafter}, despite being trained on large-scale video datasets, generate unrealistic and disjointed motion that deviates considerably from the correct actions. This reveals the limitations of large-scale pretrained video generative models in animating human actions. \emph{DreamVideo}, \emph{MotionDirector} and \emph{LAMP} finetune the models on a small set of videos containing the target actions. However, DreamVideo and MotionDirector exhibit obvious deviations from the reference images, indicating their difficulties in adapting motion to different reference images. LAMP shows smooth transition from the reference image but struggles with action fidelity, as seen in its rendering of the shoot dance, which exhibits disconnected or missing limbs, and its failure to generate the chopping wood action. In contrast, \emph{\sysname{}} not only maintains smooth transition from the reference image but also realistically generates the intended actions that resemble real videos, demonstrating its effectiveness.

\vspace{0.1em}\noindent\textbf{Generalization to Diverse Images.} To assess the generalization capability of \sysname{} beyond the HAA500 dataset, we test it on images sourced from the Internet and those generated by Stable Diffusion 3 \citep{esser2024scaling}. As shown in Figure \ref{fig:results-internet-img}, \sysname{} successfully animates actors in unrealistic scenarios, such as (a) an astronaut running in place in a virtual space and (b) a cartoon character shooting a soccer ball. Additionally, \sysname{} can animate generated images, such as (c) a humanoid alien pouring water over his head and (d) two humanoid aliens hugging. The results highlight \sysname{}'s strong generalization ability across diverse reference images. The animated videos from different methods are on the Webpage.

\begin{table}[!tb]
\setlength{\abovecaptionskip}{0.5em}
\caption{\footnotesize Quantitative comparison of different methods. The best and second-best results are \textbf{bolded} and \underline{underlined}.}
\label{tab:main}
\centering
\scriptsize
\setlength{\tabcolsep}{1.5pt}
\begin{tabular}{@{}lp{0.5pt}cccccc@{}}
    \toprule
    Method && ~~\makecell{Cosine\\RGB\\($\uparrow$)}~~ & ~~\makecell{Cosine\\Flow\\($\uparrow$)}~~ & ~\makecell{CD-FVD\\($\downarrow$)}~ & \makecell{Text\\Alignment\\($\uparrow$)} & \makecell{Image\\Alignment\\($\uparrow$)} & \makecell{Temporal\\Consistency\\($\uparrow$)} \\
    \midrule
    TI2V-Zero && 68.91 & 50.07 & 1524.08 & \underline{23.32} & 67.40 & 87.83 \\
    SparseCtrl && 67.18 & 57.16 & 1584.19 & 21.74 & 61.01 & 88.28 \\
    PIA && 69.91 & 60.04 & 1571.87 & 23.11 & 64.59 & 93.87 \\
    DynamiCrafter && 78.82 & 64.09 & 1419.68 & 23.08 & \textbf{81.75} & 95.31 \\
    DreamVideo && 67.71 & 63.40 & \underline{886.70} & \textbf{23.93} & 65.93 & 93.50 \\
    MotionDirector && 74.83 & 68.28 & 1099.32 & 21.91 & 74.82 & \underline{95.53} \\
    LAMP && \underline{83.17} & \underline{70.60} & 1240.89 & 23.17 & 78.57 & 93.81 \\
    \sysname{} && \textbf{85.48} & \textbf{77.42} & \textbf{815.11} & 23.21 & \underline{79.66} & \textbf{95.81} \\
    \bottomrule
\end{tabular}
\vspace{-0.5em}
\end{table}

\vspace{0.1em}\noindent\textbf{Automatic Evaluation Results.} 
Following \cite{wu2023tune,wu2023lamp,henschel2024streamingt2v}, we use three metrics based on CLIP \cite{radford2021learning}: \emph{Text Alignment}, \emph{Image Alignment} and \emph{Temporal Consistency}, where higher scores indicate better performance. However, CLIP may have limited ability to capture fine-grained visual details \cite{tong2024eyes}, which may affect the accuracy of these metrics. To compare generated and real videos, we utilize Fr\'{e}chet distance following \cite{xing2023dynamicrafter}. Specifically, we adopt \emph{CD-FVD} \citep{ge2024content}, which mitigates the content bias in the commonly used FVD \citep{unterthiner2018towards}, providing a better reflection of motion quality. A lower CD-FVD indicates better performance. However, since the distribution is estimated from a limited number of testing videos, CD-FVD may not fully capture the accurate distances. To provide more accurate similarity measurements between a generated video and its real counterpart with the same reference frame, we compute the cosine similarity between each generated video and its corresponding ground-truth video using the same reference image in HAA500. We calculate two metrics, \emph{Cosine RGB} and \emph{Cosine Flow}, using RGB frames and optical flows, respectively, where higher similarity values indicate better performance. For all metrics, we report the average results across all test videos. More details are described in Appendix \ref{appendix:metrics}.


Table \ref{tab:main} presents the quantitative comparison across six metrics, where \sysname{} achieves the best performance except in Text Alignment and Image Alignment. This indicates that \sysname{} excels in generating actions with the high temporal consistency and similarity to real action videos. For Text Alignment, TI2V-Zero and DreamVideo outperform \sysname{}, but both score significantly lower on Image Alignment, because they only generate text-aligned content but struggle to transition from reference images (see Figure \ref{fig:results-cmp}). For Image Alignment, DynamiCrafter surpasses \sysname{} but performs considerably worse on other metrics, as it tends to replicate the reference images rather than generate realistic actions (see Figure \ref{fig:results-cmp}).

\begin{figure*}[!tb]
\setlength{\abovecaptionskip}{0.5em}
\centering
\includegraphics[width=\linewidth]{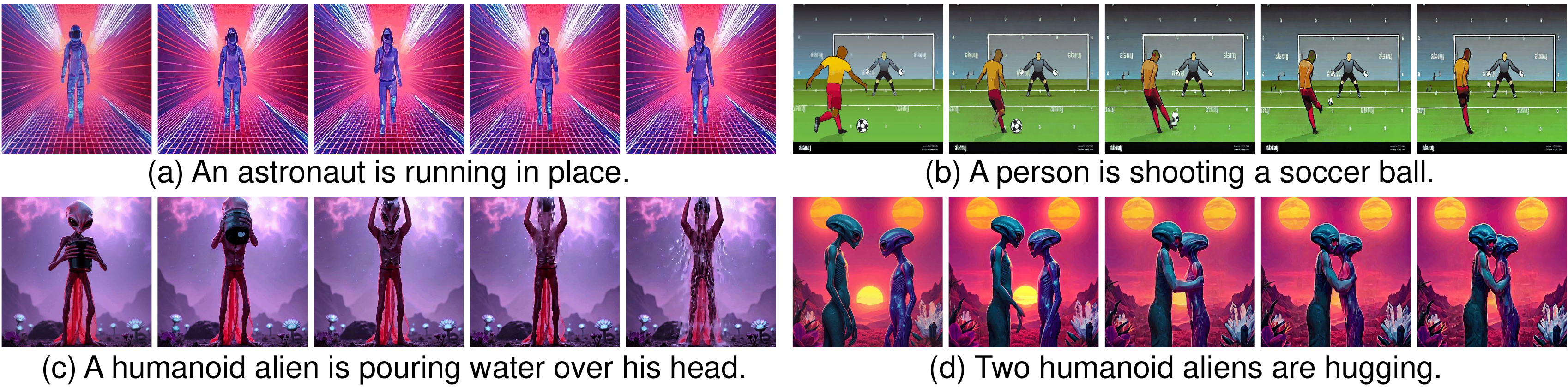}
\caption{Animated videos from \sysname{} using reference images from the Internet and generated by Stable Diffusion 3.}
\label{fig:results-internet-img}
\end{figure*}


\vspace{0.1em}\noindent\textbf{Performance on Common Actions.} Though we mainly focus on uncommon actions, our experiments also include common actions. On \textsf{\small drinking from a cup} and \textsf{\small hugging human}, \sysname{} outperforms baselines by over 100 on CD-FVD and 7\% on Cosine Flow, with comparable results on other metrics, suggesting that \sysname{} can also improve the animation quality of common actions through few-shot fintuning.

\begin{table*}[!tb]
\setlength{\abovecaptionskip}{0.2em}
\caption{Quantitative ablation studies on different components of \sysname{}. The best and second-best results are \textbf{bolded} and \underline{underlined}.}
\label{tab:ablation_component}
\footnotesize
\centering
\setlength{\tabcolsep}{3.0pt}
\setlength{\aboverulesep}{1pt}
\setlength{\belowrulesep}{1pt}
\setlength{\extrarowheight}{0pt}
\begin{tabular}{@{}cp{0.5pt}ccccp{0.5pt}cccccc@{}}
    \toprule
    Variant && \makecell{Strong\\Augmentation} & \makecell{Motion\\Features\\Alignment} & \makecell{Inter-frame\\Correspondence\\Alignment} & \makecell{Detail\\Enhancement\\Decoder} && ~\makecell{Cosine\\RGB\\($\uparrow$)}~ & ~\makecell{Cosine\\Flow\\($\uparrow$)}~ & ~\makecell{CD-FVD\\($\downarrow$)}~ & \makecell{Text\\Alignment\\($\uparrow$)} & \makecell{Image\\Alignment\\($\uparrow$)} & \makecell{Temporal\\Consistency\\($\uparrow$)} \\
    \midrule
    \#1 && \ding{56} & \ding{56} & \ding{56} & \ding{56} && 83.80 & 68.06 & 1023.30 & 22.53 & \textbf{77.10} & \textbf{95.43} \\
    \midrule
    \#2 && \ding{52} & \ding{56} & \ding{56} & \ding{56} && 83.98 & 70.61 & 932.92 & 22.48 & \underline{76.72} & 94.91 \\
    \#3 && \ding{52} & \ding{52} & \ding{56} & \ding{56} && 84.44 & 71.40 & 920.39 & 22.64 & 76.48 & 95.06 \\
    \#4 && \ding{52} & \ding{56} & \ding{52} & \ding{56} && 84.32 & 71.72 & 938.21 & \underline{22.70} & 76.31 & 94.84 \\
    \#5 && \ding{52} & \ding{52} & \ding{52} & \ding{56} && \underline{84.46} & \underline{72.24} & \textbf{906.31} & 22.52 & 76.35 & 95.01 \\
    \midrule
    \#6 && \ding{52} & \ding{52} & \ding{52} & \ding{52} && \textbf{84.51} & \textbf{72.33} & \underline{908.39} & \textbf{22.77} & 76.22 & \underline{95.31} \\
    \bottomrule
\end{tabular}
\vspace{-0.5em}
\end{table*}

\subsection{Ablation Studies}
Due to limited computational resources, we conducted ablation studies on four representative actions: \textsf{\small sprint run}, \textsf{\small soccer shoot}, \textsf{\small canoeing sprint}, and \textsf{\small hugging human}, which cover single-person actions, human-object interactions and human-human interactions and both small- and large-scale motions. We compare several model variants with incremental modifications, as outlined in Table \ref{tab:ablation_component}. The quantitative and qualitative results are presented in Table \ref{tab:ablation_component} and Figure \ref{fig:results-abl} in the Appendix, respectively. The animated videos are available on the Webpage.

Comparing the quantitative results of Variants \#1 and \#2, we observe that Variant \#2 improves CD-FVD, Cosine RGB, and Cosine Flow, albeit with a slight decrease in CLIP scores. Qualitative results show that Variant \#2  improves the fidelity of the generated actions. For example, in the soccer shooting action, the person’s legs tend to disappear as the action progresses in Variant \#1; however, Variant \#2 preserves the leg movements. These results suggests that using augmented videos improves the quality of motion.

Comparing the quantitative results of Variant \#2 with Variants \#3, \#4, and \#5, we find that Variants \#3, \#4, and \#5 improve CD-FVD, Cosine RGB, and Cosine Flow. Both Variants \#3 and \#4 enhance the Cosine RGB, and Cosine Flow. When combined, Variant \#5 yields further enhancements in cosine similarity and a 25-point improvements in CD-FVD, with only a slight decrease in Image Alignment. Qualitative results also indicates improved fidelity in Variants \#3, \#4, and \#5. For instance, motion in Variant \#2 appears unrealistic in both actions. In the soccer shooting action, the person's foot didn't touch the soccer ball, and the leg appears disconnected in some frames. In the canoe paddling action, the hand positions on the paddle are inconsistent across frames. However, these issues are largely mitigated in Variants \#3, \#4, and \#5. These results demonstrate the effectiveness of the Motion Alignment Module in learning accurate motion. By providing explicit guidance for learning appearance-general motion, the module directs the model toward generalizable motion, thereby improving the quality of the generated videos.

Comparing the quantitative results of Variant \#5 and Variant \#6, we observe that Variant \#6 noticeably improves Text Alignment and Temporal Consistency and slightly improves Cosine RGB and Cosine Flow, without substantially affecting CD-FVD. Qualitatively, Variant \#6 enhances some details (\eg, the soccer ball in certain frames in the soccer shooting action) and reduces noise in generated frames (see videos on the Webpage). These results suggest that the Detail Enhancement Decoder could compensate for detail loss or distortions in generated frames. Since the decoder operates on a frame-by-frame manner without considering inter-frame relations, it has minimal impact on motion patterns, leading to only slight effects on CD-FVD.

\vspace{0.2em}\noindent\textbf{More Results.} In Appendix \ref{appendix:ablation}, we demonstrate that the Motion Alignment Module improves motion quality across various few-shot settings (\ie, 8 or 4 videos per action class) and benefits from joint training across multiple action classes. Additionally, we conduct ablation studies on the hyperparameters of the Motion Alignment Module, the branches of the Detail Enhancement Module, and further evaluate the Detail Enhancement Module using DINO-V2 \cite{oquab2023dinov2}. In Appendix \ref{appendix:ucf_sports}, we show that \sysname{} surpasses the baselines on UCF Sports actions. Appendix \ref{appendix:non_human} shows the applicability of \sysname{} to natural scene motion.

\section{Conclusion}
We tackle the challenge of few-shot human action animation and propose \sysname{}. We introduce the Motion Alignment Module to learn generalizable motion by forcing the model to reconstruct two videos with identical motion but different appearances using the same aligned motion patterns. Additionally, we employ the Detail Enhancement Decoder to enhance transition smoothness through multi-scale detail propagation. Experiments validate the effectiveness of \sysname{} in animating diverse images.

{
    \small
    \bibliographystyle{ieeenat_fullname}
    \bibliography{main}
}

\clearpage
\renewcommand{\thesection}{A\arabic{section}}
\renewcommand{\theHsection}{A\arabic{section}}
\setcounter{section}{0}
\renewcommand{\thefigure}{A\arabic{figure}}
\setcounter{figure}{0}
\renewcommand{\thetable}{A\arabic{table}}
\setcounter{table}{0}
\setcounter{footnote}{0}
\maketitlesupplementary

\noindent
The Appendix is structured as follows:
\begin{itemize}
    \item Section \ref{appendix:comp-ai-tools} includes supplementary examples comparing videos generated by commercial AI video generators.
    \item Section \ref{appendix:method} elaborates on the details of \sysname{}.
    \item Section \ref{appendix:exp} describes detailed experimental setups.
    \item Section \ref{appendix:res} provides more experimental results.
    \item Section \ref{appendix:limitation} discusses the limitations of \sysname{}.
    \item Section \ref{appendix:ethics} presents the Ethical Statement.
\end{itemize}

\section{Comparison of videos generated by commercial AI video generators}
\label{appendix:comp-ai-tools}
\begin{figure*}[!ht]
\centering
\includegraphics[width=\linewidth]{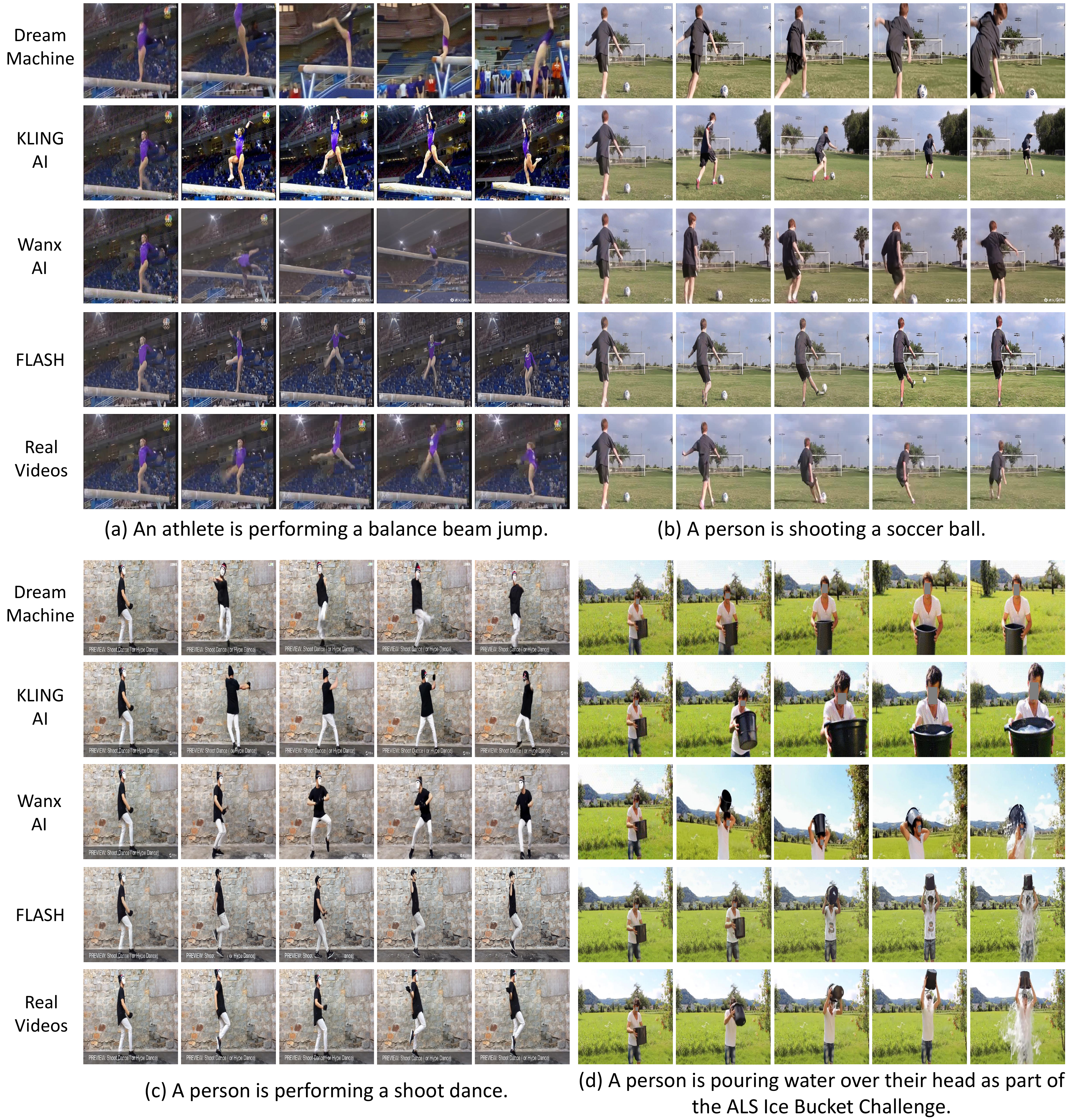}
\caption{Comparison of human action videos generated by Dream Machine, KLING AI, Wanx AI and \sysname{} (our method). Human faces are anonymized for privacy protection.}
\label{fig:comparing-tools-2}
\end{figure*}

In Figure \ref{fig:comparing-tools-2}, we present four examples of animated human action videos from Dream Machine\footnote{\url{https://lumalabs.ai/dream-machine}}, KLING AI\footnote{\url{https://www.klingai.com/}}, Wanx AI\footnote{\url{https://tongyi.aliyun.com/wanxiang/}}, and \sysname{}. The videos are available on the Webpage. Dream Machine, KLING AI and Wanx AI struggle to animate these actions accurately. In the balance beam jump action, Dream Machine and Wanx AI produce unrealistic, physics-defying movements, while KLING AI generates a jump but fails to depict standard jumps on the balance beam. For the soccer shooting action, all three models fail to generate a correct shooting motion, with the person never kicking the ball. In the shoot dance action, Dream Machine and KLING AI generate unnatural, physically implausible movements, whereas Wanx AI produces dance movements but does not capture the shoot dance correctly. In the Ice Bucket Challenge action, none of the three models accurately portray the motion of pouring ice water from the bucket onto the body. In contrast, \sysname{} generates these actions with higher fidelity to the real actions.

\section{\sysname{}}
\label{appendix:method}

\subsection{Components in Latent Video Diffusion Models}
\label{appendix:pre}
\textbf{Temporal Attention Layers.} To capture temporal dynamics in videos, \cite{ho2022video,esser2023structure,guo2023animatediff,guo2023sparsectrl} add temporal attention layers after each spatial attention layer of U-Net. In each temporal attention layer, we first reshape the input features $\boldsymbol{F}_{in} \in \mathbb{R}^{N \times h' \times w' \times c'}$ to $\tilde{\boldsymbol{F}}_{in} \in \mathbb{R}^{B \times N \times c'}$, where $B = h' \times w'$. Here, we treat the features at different spatial locations as independent samples. Then, we add temporal position encoding to $\tilde{\boldsymbol{F}}_{in}$ and employ a self-attention layer to transform $\tilde{\boldsymbol{F}}_{in}$ into $\tilde{\boldsymbol{F}}_{out} \in \mathbb{R}^{B \times N \times c'}$. Finally, we reshape $\tilde{\boldsymbol{F}}_{out}$ to $\boldsymbol{F}_{out}\in\mathbb{R}^{N \times h' \times w' \times c'}$ as the output features. The temporal attention layer integrates information from different frames for each spatial location, enabling the learning of temporal changes.

\vspace{0.2em}\noindent\textbf{Cross-frame Attention Layers.} To enhance temporal consistency across generated frames, \cite{text2video-zero,wu2023lamp} replace spatial self-attention layers with spatial cross-frame attention layers. While self-attention layers use features from the current frame as key and value, cross-frame attention layers restrict key and value to the features from the first frame. These layers carry over the appearance features from the first frame to subsequent frames, improving temporal consistency in the generated videos.

\vspace{0.2em}\noindent\textbf{Noise-Free Frame Conditioning.} To further preserve the appearance of the reference image in the image animation task, \cite{wu2023lamp,ren2024consisti2v} keep the latent reference image noise-free in the noised latent video. Specifically, at the noising step $t$, the latent video $\boldsymbol{Z}_t=\langle\boldsymbol{z}_t^i\rangle_{i=1}^N$ is modified to $\check{\boldsymbol{Z}}_t=\langle\boldsymbol{z}_0^1, \boldsymbol{z}_t^2, \cdots, \boldsymbol{z}_t^N\rangle$, where $\boldsymbol{z}_t^1$ is replaced by $\boldsymbol{z}_0^1$, which is noise-free. During inference, a sample $\boldsymbol{Z}_T$ is drawn from $\mathcal{N}(\boldsymbol{0},\,I)$, and $\boldsymbol{z}_T^1$ is substituted with $\boldsymbol{z}_0^1=\mathcal{E}(I)$, where $I$ is the user-provided reference image. The modified latent video $\check{\boldsymbol{Z}}_T=\langle\boldsymbol{z}_0^1, \boldsymbol{z}_T^2, \cdots, \boldsymbol{z}_T^N\rangle$ is then used for denoising. This technique effectively maintain the features from the first frame in subsequent frames.

\sysname{} adopts these components in its base video diffusion model, and designs the Motion Alignment Module and the Detail Enhancement Decoder on top of it.

\begin{figure*}[tb]
\centering
\includegraphics[width=\linewidth]{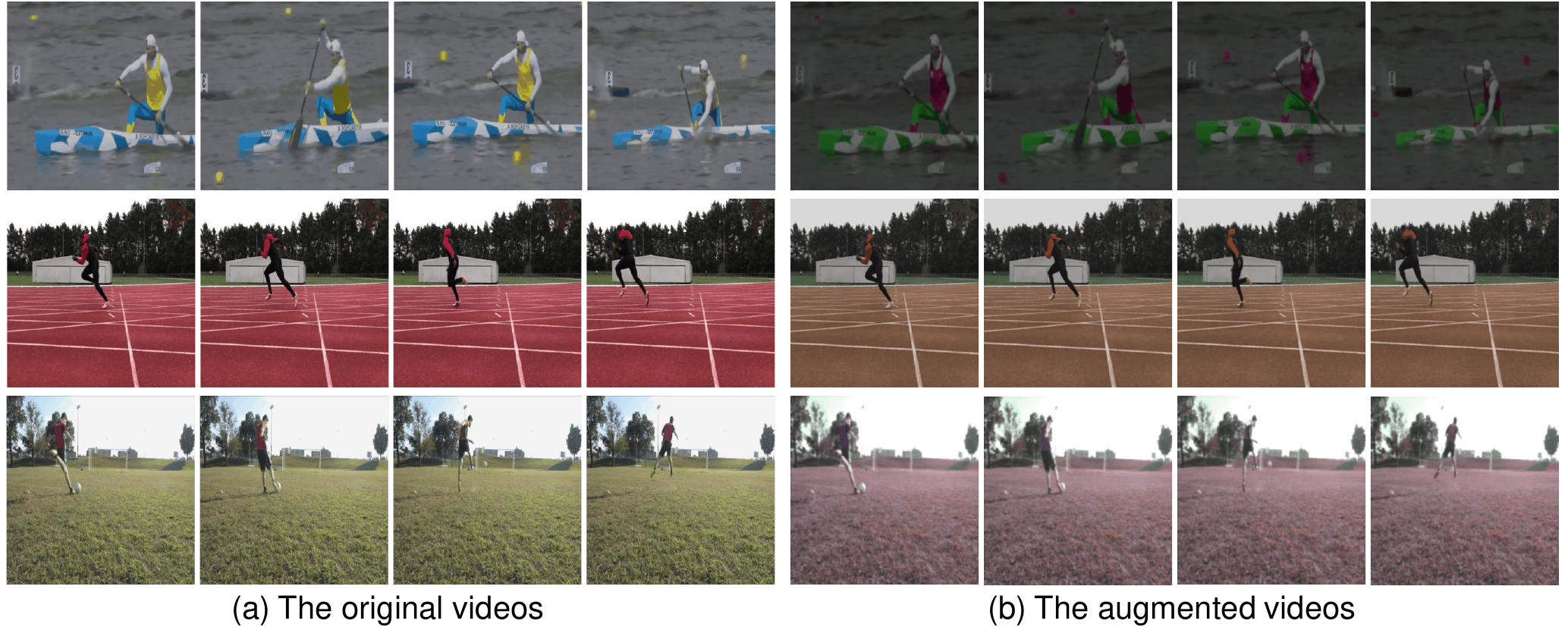}
\caption{Examples of three original videos alongside their corresponding strongly augmented videos.}
\label{fig:aug_videos}
\end{figure*}

\subsection{Strongly Augmented Videos}
\label{appendix:strong_aug_video}
To create a strongly augmented version of an original video, we sequentially apply Gaussian blur and random color adjustments to the original video. This process is designed to preserve the original motion while altering the appearance uniformly across all frames.

\begin{itemize}
\item Gaussian blur: A kernel size is randomly selected from a predefined range, as specified in Sec. \ref{appendix:imple_detail}. This kernel size is used to apply Gaussian blur to every frame of the original video, ensuring a uniform level of blur throughout.
\item Random color adjustments: After applying Gaussian blur, we randomly adjust the brightness, contrast, saturation, and hue of the video. For each property, an adjustment factor is randomly chosen from its respective predefined range, detailed in Sec. \ref{appendix:imple_detail}. The adjustment with the chosen factor is applied uniformly across all frames to maintain consistent color alterations without introducing cross-frame inconsistencies. We implement it with the \emph{ColorJitter} function in PyTorch.
\end{itemize}

By applying these augmentations with consistent parameters across all frames, the augmented video retains the motion in the original video while showing altered appearances. Figure \ref{fig:aug_videos} presents examples of strongly augmented videos. These augmented videos exhibit considerable differences from the original ones in aspects such as the background and the actors' clothing. However, the motion from the original videos is preserved.

\subsection{Detail Enhancement Decoder}
\label{appendix:decoder}
\noindent\textbf{Multi-scale Detail Propagation.} Before feeding the two features $\boldsymbol{g}_{l}^1$ and $\boldsymbol{h}_{l}^i$ to the two branches, we first interpolate $\boldsymbol{g}_{l}^1$ to match the spatial size of $\boldsymbol{h}_{l}^i$ and use a fully connected layer to adjust $\boldsymbol{g}_{l}^1$ to the same number of channels as $\boldsymbol{h}_{l}^i$, resulting $\tilde{\boldsymbol{g}}_k^1$ as the input of the two branches. The network $\mathcal{N}$ in the \emph{Warping Branch} is a four-layer convolution network that takes the channel-wise concatenation of $\boldsymbol{h}_l^i$ and $\tilde{\boldsymbol{g}}_l^1$ as input and outputs the spatial displacements $(\Delta p, \Delta q)$. In the \emph{Patch Attention Branch}, before applying the cross-attention layer $\mathcal{A}$, we use a fully connected layer to transform each patch of the two features into a feature vector. The network $\mathcal{M}$, which generates the fusion weights, is a two-layer convolution network, which takes the channel-wise concatenation of $\boldsymbol{h}_l^i$ and $\tilde{\boldsymbol{g}}_l^1$ as input and outputs the fusion weights $\boldsymbol{w}^i_l$.

\noindent\textbf{Distorted Videos.} The details of video distortions are as follows: The random Gaussian blur and random color adjustments follow the implementations described in Sec. \ref{appendix:strong_aug_video}. However, the random color adjustments here differ in that they are applied to only 80\% of randomly selected regions rather than to all regions. This modification is intentional, as the goal is to create distorted videos with inconsistent color changes that simulate the distortions in latent videos, rather than to maintain consistent color changes as in Sec. \ref{appendix:strong_aug_video}. For random elastic transformations, displacement vectors are generated for all spatial positions based on random offsets sampled from a predefined range (detailed in Sec. \ref{appendix:imple_detail}) and are then used to transform each pixel accordingly. We implement it using the \emph{ElasticTransform} function in PyTorch.

\section{Experiment Details}
\label{appendix:exp}
\subsection{Data}
\label{appendix:data}
We conduct experiments on 16 actions selected from the HAA500 dataset \citep{chung2021haa500}, which contains 500 human-centric atomic actions capturing the precise movements of human, each consisting of 20 short videos. The selected actions include single-person actions (\textsf{\small push-up}, \textsf{\small arm wave}, \textsf{\small shoot dance}, \textsf{\small running in place}, \textsf{\small sprint run}, and \textsf{\small backflip}), human-object interactions (\textsf{\small soccer shoot}, \textsf{\small drinking from a cup}, \textsf{\small balance beam jump}, \textsf{\small balance beam spin}, \textsf{\small canoeing sprint}, \textsf{\small chopping wood}, \textsf{\small ice bucket challenge}, and \textsf{\small basketball hook shot}), and human-human interactions (\textsf{\small hugging human}, \textsf{\small face slapping}).

\vspace{0.5em}\noindent\textbf{Training videos.} For each selected action, we use 16 videos from the training split in HAA500 for training. We manually exclude videos that contain pauses or annotated symbols in the frames. Each action label is converted into a natural sentence as the action description; for example, the action label ``soccer shoot'' is converted to ``a person is shooting a soccer ball.''

\begin{figure}[tb]
\centering
\includegraphics[width=\linewidth]{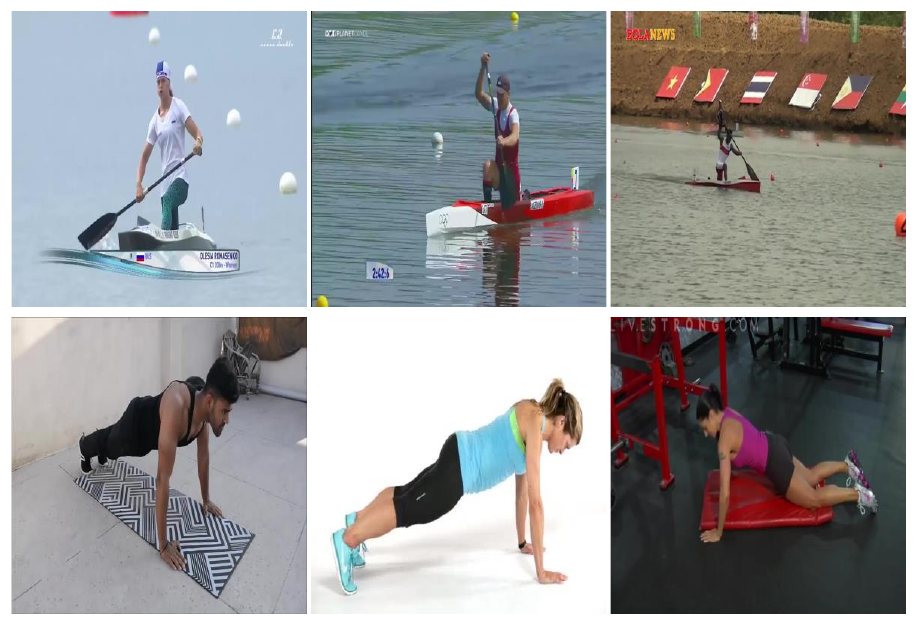}
\caption{Similarity between training videos in the same action class. The first row presents three videos depicting the action \textsf{\small canoeing sprint}, and the second row showcases three videos containing the action \textsf{\small push-up}.}
\label{fig:intra_class_sim}
\end{figure}

\vspace{0.5em}\noindent\textbf{Similarity between training videos in the same action class.} Videos within the same action class do not share similar visual characteristics, such as scenes, viewing angles, actor positions, or shot types (\eg, close-up or wide shot), as shown in the examples in Figure \ref{fig:intra_class_sim}.

\vspace{0.5em}\noindent\textbf{Testing images.} For each selected action, we use the first frames from the four testing videos as testing images. Additionally, we search online for two human images depicting a person beginning the desired action as additional testing images.

\subsection{Implementation Details}
\label{appendix:imple_detail}
We use AnimteDiff \citep{he2023animate} as the base video generative model. We initialize all parameters with pretrained weights of AnimteDiff. The spatial resolution of generated videos is set to $512 \times 512$, and the video length is set to 16 frames.

\vspace{0.5em}\noindent\textbf{Training of U-Net.} We combine features from the first and current frames as keys and values in the spatial cross-frame attention layers. Following \cite{huang2023reversion,materzynska2023customizing}, we redefine the sampling probability distribution to prioritize earlier denoising stages. In the Motion Alignment Module, we set $\tau$ to 90 and apply motion feature alignment after each temporal attention layer in the U-Net. Inter-frame correspondence alignment is applied to 50\% of the cross-frame attention layers, selected randomly. For simplicity, we replace $\boldsymbol{Q}$ and $\boldsymbol{K}$ of the augmented video with those of the original video when calculating $S$, instead of directly replacing $S$. Gaussian blur is applied with a randomly sampled kernel size between 3 and 10. Random color adjustment modifies brightness, saturation, and contrast by random factors between 0.5 and 1.5, and modifies hue by a random factor between -0.25 and 0.25. Before applying strong augmentations to the original video, we first perform random horizontal flipping and random cropping on the original video. We only train the temporal attention layers, and the key and value matrices of spatial attention layers. The learning rate is set to $5.0\times 10^{-5}$, with training conducted for 20,000 steps.

\vspace{0.5em}\noindent\textbf{Training of Detail Enhancement Decoder.} The patch size in the Patch Attention Branch is set to 2. For video distortion, Gaussian blur is applied with a random kernel size between 3 and 10. Random color adjustment use random factors for brightness, saturation, and contrast between 0.7 and 1.3, and a random factor for hue between -0.2 and 0.2. For random elastic transformations, displacement strength is randomly sampled from 1 to 20. We only train the newly added layers, with a learning rate of $1.0\times 10^{-4}$ over 10,000 steps.

\vspace{0.5em}\noindent\textbf{Inference.} During inference, we utilize the DDIM sampling process \citep{song2020denoising} with 25 denoising steps. Classifier-free guidance \citep{ho2022classifier} is applied with a guidance scale set to 7.5. Following \cite{wu2023lamp}, we apply AdaIN \citep{huang2017arbitrary} on latent videos for post-processing.

\vspace{0.5em}\noindent\textbf{Computational Resources.} Our experiments are conducted on a single GeForce RTX 3090 GPU using PyTorch, with a batch size of 1 on each GPU. We build upon the codebase of AnimateDiff \citep{guo2023animatediff}. Training takes approximately 36 hours per action.

\begin{figure*}[tb]
\centering
\includegraphics[width=\linewidth]{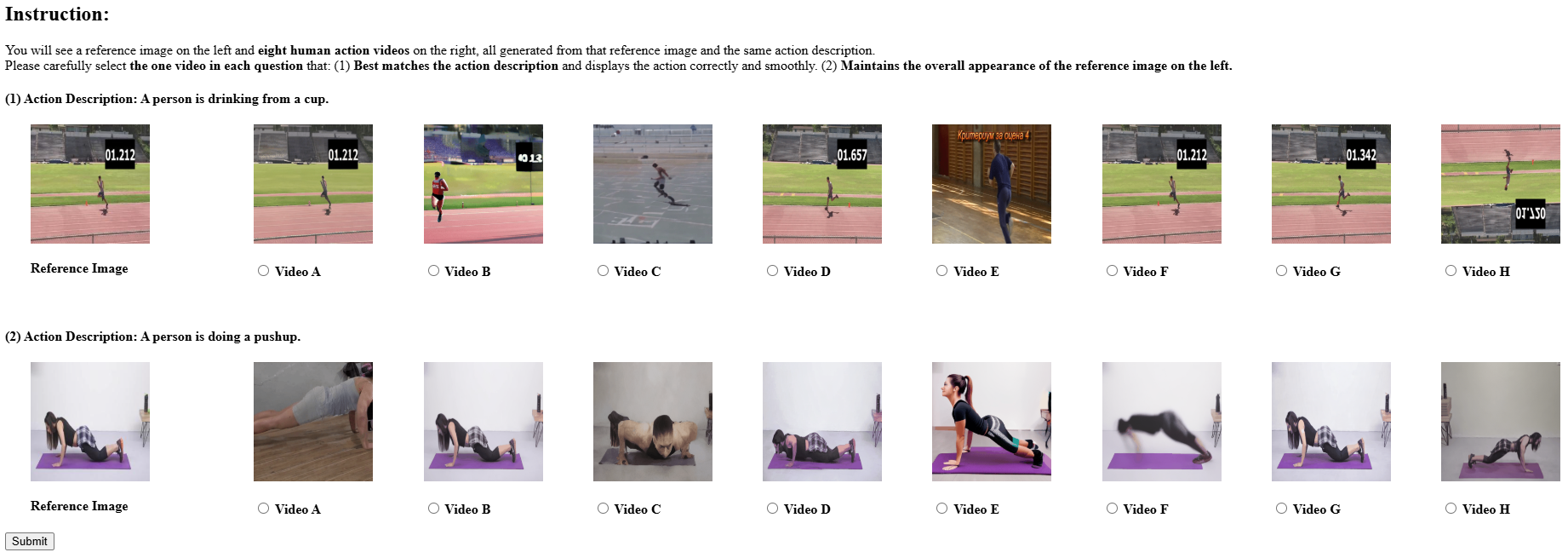}
\caption{AMT user study interface.}
\label{fig:AMT_interface}
\end{figure*}


\begin{figure*}[!tb]
\centering
\includegraphics[width=\linewidth]{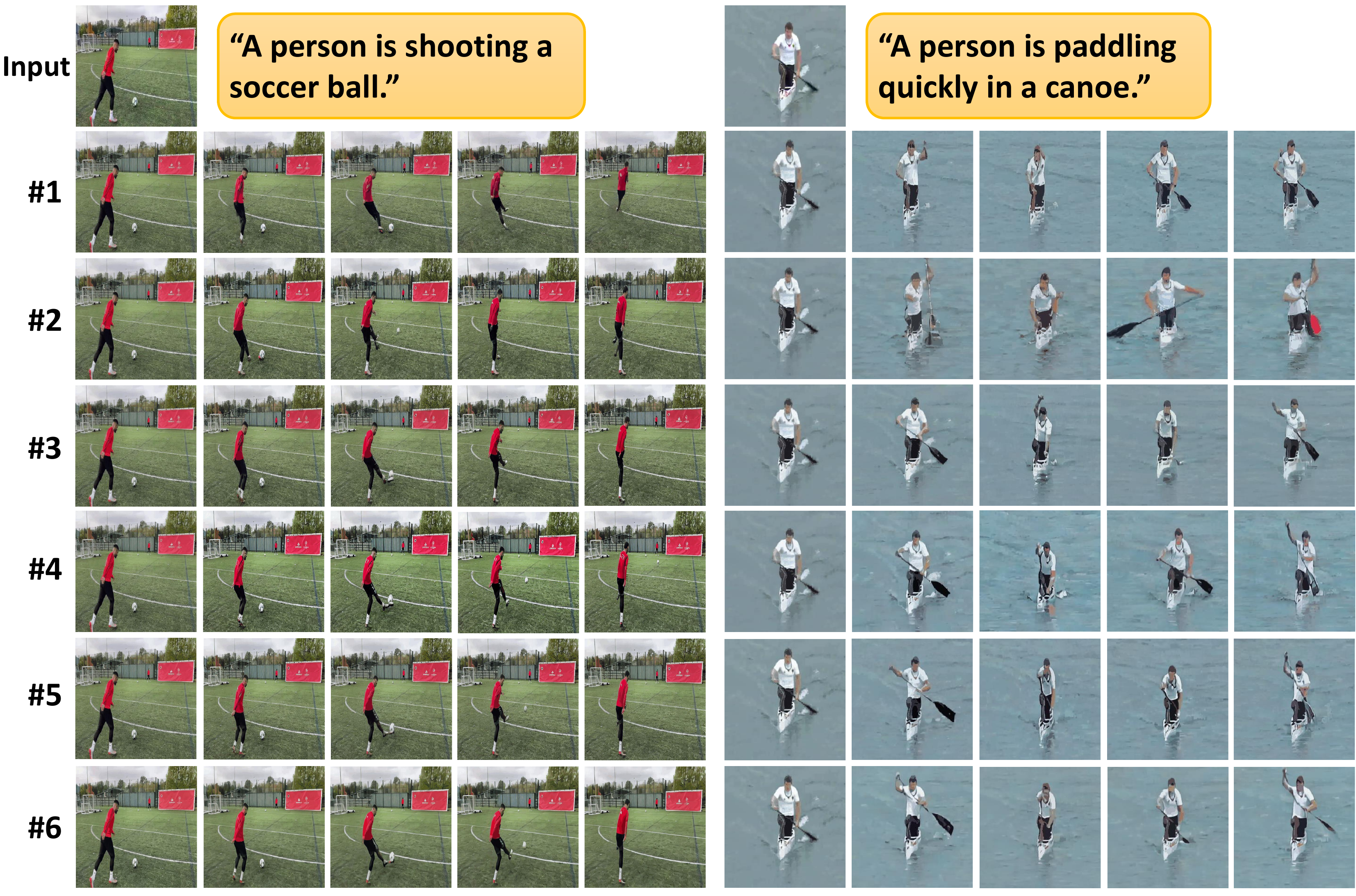}
\caption{Qualitative ablation study on different components of \sysname{}. \#1: baseline model, \#2: model trained with strongly augmented videos, \#3: with motion feature alignment, \#4: with inter-frame correspondence alignment, \#5: with both alignments, \#6: full model with the Motion Alignment Module and the Detail Enhancement Decoder. See Table \ref{tab:ablation_component} for the details of each variant.}
\label{fig:results-abl}
\end{figure*}

\subsection{Evaluation Metrics}
\label{appendix:metrics}
In line with previous works \citep{wu2023tune, wu2023lamp, henschel2024streamingt2v}, we use three metrics based on CLIP \cite{radford2021learning} to assess text alignment, image alignment, and temporal consistency. (1) \emph{Text Alignment}: We compute the similarity between the visual features of each frame and the textual features of the text prompt, and average the similarities across all frames. (2) \emph{Image Alignment}: We compute the similarity between the visual features of each frame and the visual features of the provided reference image, and average the similarities across all frames. (3) \emph{Temporal Consistency}: We calculate the average similarity between the visual features of consecutive frame pairs to obtain the temporal consistency score. We use ViT-L/14 from OpenAI \citep{radford2021learning} for feature extraction. In these three metrics, higher scores indicate better performance.

Following \cite{xing2023dynamicrafter}, we utilize Fr\'{e}chet distance to compare generated and real videos. We use \emph{CD-FVD} \citep{ge2024content} to mitigate content bias in the widely used FVD \citep{unterthiner2018towards}. We use VideoMAE \citep{tong2022videomae}, pretrained on SomethingSomethingV2 \citep{goyal2017something}, for feature extraction and calculate distance between real and generated videos. In this metric, lower distances indicate better performance.

To evaluate the similarity between generated videos and ground-truth videos in the HAA dataset, we calculate the cosine similarity for each pair of the generated and ground-truth videos. (1) \emph{Cosine RGB}: We extract video features using I3D \citep{carreira2017quo}, pretrained on RGB videos, for both the generated and ground truth videos, and calculate cosine similarity for the pair. (2) \emph{Cosine Flow}: We extract optical flow using RAFT \citep{teed2020raft} and then use I3D \citep{carreira2017quo}, pretrained on optical flow data, to extract video features for cosine similarity calculation. In these two metrics, higher similarities indicate better performance.

\subsection{Baselines}
\label{appendix:baselines}
We compare \sysname{} with several baselines: (1) TI2V-Zero \citep{ni2024ti2v}, a training-free image animation model that injects the appearance of the reference image into a pretrained text-to-video model. We directly test image animation on its checkpoints. (2) SparseCtrl \citep{guo2023sparsectrl}, an image animation model that encodes the reference image with a sparse condition encoder and integrates the features into a text-to-video model. It is trained on large-scale video datasets, and we directly test image animation on its checkpoints. (3) PIA \citep{zhang2023pia}, an image animation model that incorporates the reference image features into the noised latent video. It is trained on large-scale video datasets, and we directly test image animation on its checkpoints. (4) DynamiCrafter \citep{xing2023dynamicrafter}, an image animation model that injects the reference image features into generated videos via cross-attention layers and feature concatenation. It is trained on large-scale video datasets, and we directly test image animation on its checkpoints. (5) DreamVideo \citep{wei2024dreamvideo}, a customized video generation model which learns target subject and  motion using a limited set of samples. We train it to customize motion for each action using the same training videos as \sysname{}. (6) MotionDirector \cite{zhao2024motiondirector}, a customized video generation model which learns target appearance and motion with limited videos. We train its motion adapter with the same training videos as \sysname{} and its appearance adapter with the testing reference images. Therefore, MotionDirector has access to more data (the testing reference images) than other methods. (7) LAMP \citep{wu2023lamp}, a few-shot image animation model which learns motion patterns from a few videos. We train it with the same training videos as \sysname{}.

\begin{table*}[!tb]
\caption{Analysis of training with fewer videos and joint training with multiple action classes.}
\label{tab:ablation_settings}
\centering
\begin{tabular}{@{}cp{1pt}ccp{1pt}cccccc@{}}
    \toprule
    Variant && \makecell{\# Videos\\Per Class} & \makecell{joint\\Training} && ~\makecell{Cosine\\RGB\\($\uparrow$)}~ & ~\makecell{Cosine\\Flow\\($\uparrow$)}~ & ~\makecell{CD-FVD\\($\downarrow$)}~ & \makecell{Text\\Alignment\\($\uparrow$)} & \makecell{Image\\Alignment\\($\uparrow$)} & \makecell{Temporal\\Consistency\\($\uparrow$)} \\
    \midrule
    \#1 && 16 & \ding{56} && 83.80 & 68.06 & 1023.30 & 22.53 & 77.10 & 95.43 \\
    \#2 && 16 & \ding{56} && 83.98 & 70.61 & 932.92 & 22.48 & 76.72 & 94.91 \\
    \#5 && 16 & \ding{56} && 84.46 & 72.24 & 906.31 & 22.52 & 76.35 & 95.01 \\
    \midrule
    \#1 && 8 & \ding{56} && 82.50 & 68.13 & 995.43 & 22.70 & 76.05 & 94.79 \\
    \#2 && 8 & \ding{56} && 83.30 & 70.09 & 962.82 & 22.62 & 74.37 & 94.40 \\
    \#5 && 8 & \ding{56} && 83.40 & 72.01 & 943.54 & 22.66 & 75.02 & 94.51 \\
    \midrule
    \#1 && 4 & \ding{56} && 81.40 & 68.02 & 1050.03 & 22.22 & 72.81 & 94.24 \\
    \#2 && 4 & \ding{56} && 81.88 & 70.15 & 1045.49 & 22.60 & 72.00 & 93.83 \\
    \#5 && 4 & \ding{56} && 82.22 & 71.83 & 1031.87 & 22.46 & 72.56 & 94.22 \\
    \midrule
    \#5 && 16 & \ding{56} && 84.46 & 72.24 & 906.31 & 22.52 & 76.35 & 95.01 \\
    \#5 && 16 & \ding{52} && 85.01 & 72.32 & 897.05 & 22.61 & 77.47 & 95.39 \\
    \bottomrule
\end{tabular}
\end{table*}

\begin{table*}[!tb]
\caption{Ablation studies on different values of $\tau$ for motion feature alignment, different values of $p$ for inter-frame correspondence alignment, and the impact of the Warping Branch and Patch Attention Branch in the Detail Enhancement Decoder.}
\label{tab:ablation_params}
\centering
\setlength{\tabcolsep}{2.0pt}
\begin{tabular}{@{}cp{1pt}ccccp{1pt}cccccc@{}}
    \toprule
    Variant && ~~~~$\tau$~~~~ & ~~~~$p$~~~~ & ~~\makecell{Warping\\Branch}~~ & ~~\makecell{Patch Attention\\Branch}~~ && ~\makecell{Cosine\\RGB\\($\uparrow$)}~ & ~\makecell{Cosine\\Flow\\($\uparrow$)}~ & ~\makecell{CD-FVD\\($\downarrow$)}~ & \makecell{Text\\Alignment\\($\uparrow$)} & \makecell{Image\\Alignment\\($\uparrow$)} & \makecell{Temporal\\Consistency\\($\uparrow$)} \\
    \midrule
    \#3 && ~90~ & - & - & - && 84.44 & 71.40 & 920.39 & 22.64 & 76.48 & 95.06 \\
    \#3 && 75 & - & - & - && 84.38 & 71.19 & 904.25 & 22.58 & 76.63 & 95.16 \\
    \#3 && 50 & - & - & - && 84.30 & 70.31 & 934.84 & 22.57 & 77.29 & 95.14 \\
    \#3 && 25 & - & - & - && 84.71 & 69.79 & 930.53 & 22.33 & 76.52 & 94.85 \\
    \midrule
    \#4 && - & ~1.0~ & - & - && 84.22 & 69.34 & 914.12 & 22.50 & 76.43 & 94.91 \\
    \#4 && - & 0.5 & - & - && 84.32 & 71.72 & 938.21 & 22.70 & 76.31 & 94.84 \\
    \midrule
    \#5 && 90 & 0.5 & \ding{56} & \ding{56} && 84.46 & 72.24 & 906.31 & 22.52 & 76.35 & 95.01 \\
    \#6 && 90 & 0.5 & \ding{52} & \ding{56} && 84.63 & 71.96 & 918.61 & 22.54 & 76.21 & 95.35 \\
    \#6 && 90 & 0.5 & \ding{56} & \ding{52} && 83.32 & 72.26 & 888.05 & 22.71 & 74.97 & 95.13 \\
    \#6 && 90 & 0.5 & \ding{52} & \ding{52} && 84.51 & 72.33 & 908.39 & 22.77 & 76.22 & 95.31 \\
    \bottomrule
\end{tabular}
\end{table*}

\begin{table}[!tb]
\caption{Evaluation of the Detail Enhancement Decoder with DINO-V2.}
\label{tab:ablation_dino}
\footnotesize
\centering
\setlength{\tabcolsep}{8.0pt}
\begin{tabular}{@{}cp{1pt}cc}
    \toprule
    Variant && \makecell{Image Alignment ($\uparrow$)} & \makecell{Temporal Consistency ($\uparrow$)} \\
    \midrule
    \#5 && 87.02 & 97.48 \\
    \#6 && 87.75 & 97.85 \\ 
    \bottomrule
\end{tabular}
\end{table}

\section{Results}
\label{appendix:res}

\subsection{User Study}
\label{appendix:user_study}
We conducted a user study on Amazon Mechanical Turk (AMT), where workers were asked to select the best generated video from a set of candidates. For each action, we randomly selected four different reference images and their corresponding generated videos for this user study. The AMT assessment interface, shown in Figure \ref{fig:AMT_interface}, presented workers with the following instructions: ``You will see a reference image on the left and eight human action videos on the right, all generated from that reference image and an action description. Please carefully select the one video in each question that: (1) Best matches the action description and displays the action correctly and smoothly. (2) Maintains the overall appearance of the reference image on the left.'' The interface also displayed the reference image and action description.

To identify random clicking, each question was paired with a control question that has obvious correct answer. The control question includes a real video of a randomly selected action alongside clearly incorrect ones, such as a static video, a video with shuffled frames, and a video from the same action class that mismatches the reference image. The main question and the control question were randomly shuffled within each question pair, and each pair was evaluated by 10 different workers. Responses from workers who failed the control questions were marked as invalid.

In total, we collected 488 valid responses. The preference rates for different methods are shown in the pie chart in Figure \ref{fig:AMT_results} in the main paper. \sysname{} was preferred in 66\% of the valid responses, significantly outperforming the next best choices, DynamiCrafter(13\%) and LAMP (11\%).

\subsection{Additional Ablation Studies}
\label{appendix:ablation}
In line with the main paper, Variant \#1 serves as the baseline, excluding both the Motion Alignment Module and the Detail Enhancement Decoder. Variant \#2 uses strongly augmented videos for training without any alignment technique. Variants \#3, \#4, and \#5 progressively incorporate motion feature alignment, inter-frame correspondence alignment, and both, respectively, on top of Variant \#2. Lastly, Variant \#6 builds upon Variant \#5 by incorporating the Detail Enhancement Decoder.

\vspace{0.5em}\noindent\textbf{Applicability with Fewer Training Videos.} To assess the few-shot learning capability of the Motion Alignment Module, we conduct experiments using 8 and 4 videos randomly sampled from each action class. The results are shown in Table \ref{tab:ablation_settings}. Across different numbers of training videos per action class, Variant \#5 consistently outperforms Variants \#1 and \#2 on CD-FVD, Cosine-RGB and Cosine-Flow. The results show that the Motion Alignment Module enhances the motion quality of animated videos in different few-shot configurations.

\vspace{0.5em}\noindent\textbf{Joint Training with Multiple Action Classes.} We examine whether the model benefits from joint training across multiple action classes. We use all the training videos from the four action classes (\textsf{\small sprint run}, \textsf{\small soccer shoot}, \textsf{\small canoeing sprint}, and \textsf{\small hugging human}) to train a single model. The results in Table \ref{tab:ablation_settings} show improvements across all metrics. The improvements in Image Alignment, Temporal Consistency, and Cosine RGB are considerable. The results suggest that joint training with multiple action classes enhances the quality of the generated videos. This makes our technique more practical for applications that have accessible example videos of multiple delicate or customized human actions.

\vspace{0.5em}\noindent\textbf{Analysis of Motion Alignment Module.} In Table \ref{tab:ablation_params}, we compare the performance of different $\tau$ values in Variant \#3 and different $p$ values in Variant \#4. For $\tau$, we observe that decreasing $\tau$ reduces performance in Temporal Consistency, CD-FVD, and Cosine Flow, especially in Temporal Consistency (94.85 for $\tau=25$) and Cosine Flow (69.79 for $\tau=25$). This suggests that including more channels as motion channels degrades video quality, likely because motion information is only encoded in a limited number of channels \citep{xiao2024video}, and aligning too many channels hampers feature learning. Thus, we set $\tau=90$ for the remaining experiments. Regarding $p$, substituting inter-frame correspondence relations in all cross-frame attention layers ($p=1.0$) lowers Cosine Flow significantly (\eg, 69.34 for $p=1.0$) but doesn't affect other metrics obviously. This might be due to the excessive regularization from substituting inter-frame correspondence relations in every layer, which makes learning difficult. Therefore, we use $p=0.5$ in the remaining experiments.

\vspace{0.5em}\noindent\textbf{Analysis of Detail Enhancement Decoder.} In Table \ref{tab:ablation_params}, we compare the effects of the Warping Branch and the Patch Attention Branch in Variant \#6. Using only the Warping Branch leads to a notable improvement in Temporal Consistency (from 95.01 to 95.35). In contrast, the Patch Attention Branch provides a modest increase in Text Alignment (from 22.52 to 22.71) but results in a significant drop in Image Alignment (from 76.35 to 74.97). When both branches are combined, there is an enhancement in both Text Alignment and Temporal Consistency, accompanied by only a slight decrease in Image Alignment. These results suggest that the two branches have complementary effects. Therefore, we use the two branches in the Detail Enhancement Decoder.

\vspace{0.5em}\noindent\textbf{Evaluation of the Detail Enhancement Decoder with DINO.} The CLIP vision encoder, trained on vision-language tasks, may have limited ability to perceive fine-grained visual details \cite{tong2024eyes}, which can affect the evaluation of Image Alignment and Temporal Consistency. Therefore, we use the DINO-V2 \cite{oquab2023dinov2} vision encoder, which excels at capturing rich, fine-grained details at the pixel level, to assess Image Alignment and Temporal Consistency. The results in Table \ref{tab:ablation_dino} demonstrate that the Detail Enhancement Decoder enhances both Image Alignment and Temporal Consistency, illustrating its effectiveness in improving transition smoothness.

\begin{table}[!tb]
\caption{Quantitative comparison of different methods on the UCF Sports Action Dataset. The best and second-best results are \textbf{bolded} and \underline{underlined}.}
\label{tab:main-ucf-sports}
\centering
\scriptsize
\setlength{\tabcolsep}{1.5pt}
\begin{tabular}{@{}lp{0.5pt}cccccc@{}}
    \toprule
    Method && ~~\makecell{Cosine\\RGB\\($\uparrow$)}~~ & ~~\makecell{Cosine\\Flow\\($\uparrow$)}~~ & ~\makecell{CD-FVD\\($\downarrow$)}~ & \makecell{Text\\Alignment\\($\uparrow$)} & \makecell{Image\\Alignment\\($\uparrow$)} & \makecell{Temporal\\Consistency\\($\uparrow$)} \\
    \midrule
    TI2V-Zero && 71.90 & 64.43 & 1222.35 & \underline{24.62} & 70.16 & 88.87 \\
    SparseCtrl && 71.56 & 63.21 & 1574.69 & 23.26 & 61.16 & 89.69 \\
    PIA && 70.05 & 58.51 & 1385.54 & 23.93 & 66.03 & 94.41 \\
    DynamiCrafter && \underline{77.83} & 63.16 & 1630.83 & 23.95 & \textbf{87.84} & \underline{96.75} \\
    DreamVideo && 68.60 & 70.20 & \underline{949.72} & \textbf{26.04} & 78.20 & 96.19 \\
    MotionDirector && 75.60 & 63.01 & 1315.36 & 23.88 & 76.92 & \textbf{97.07} \\
    LAMP && 74.15 & \underline{73.78} & 1076.77 & 24.02 & 81.17 & 95.17 \\
    \sysname{} && \textbf{86.80} & \textbf{79.36} & \textbf{480.70} & 24.11 & \underline{85.75} & 96.22 \\
    \bottomrule
\end{tabular}
\end{table}

\begin{table}[!tb]
\caption{Quantitative comparison of different methods on non-human motion videos. The best and second-best results are \textbf{bolded} and \underline{underlined}.}
\label{tab:main-non-human}
\centering
\scriptsize
\setlength{\tabcolsep}{1.5pt}
\begin{tabular}{@{}lp{0.5pt}cccccc@{}}
    \toprule
    Method && ~~\makecell{Cosine\\RGB\\($\uparrow$)}~~ & ~~\makecell{Cosine\\Flow\\($\uparrow$)}~~ & ~\makecell{CD-FVD\\($\downarrow$)}~ & \makecell{Text\\Alignment\\($\uparrow$)} & \makecell{Image\\Alignment\\($\uparrow$)} & \makecell{Temporal\\Consistency\\($\uparrow$)} \\
    \midrule
    TI2V-Zero && 58.96 & 45.31 & 1562.76 & 21.95 & 79.05 & 93.00 \\
    SparseCtrl && 67.89 & 59.02 & 1441.31 & 21.92 & 76.89 & 93.75 \\
    PIA && 68.11 & 57.02 & 1591.11 & 21.83 & 79.44 & 96.83 \\
    DynamiCrafter && 77.14 & 69.90 & 1371.39 & \underline{22.18} & \textbf{87.27} & \textbf{98.08} \\
    DreamVideo && 68.80 & 61.44 & 1222.22 & \textbf{22.75} & 84.40 & 96.96 \\
    MotionDirector && 74.57 & 68.41 & 1302.02 & 20.75 & 79.09 & 96.68 \\
    LAMP && \underline{79.48} & \underline{71.89} & \underline{1210.55} & 22.14 & \underline{86.68} & 97.49 \\
    \sysname{} && \textbf{79.53} & \textbf{75.48} & \textbf{1204.72} & 22.05 & 85.42 & \underline{97.51} \\
    \bottomrule
\end{tabular}
\end{table}

\begin{figure*}[tb]
\centering
\includegraphics[width=\linewidth]{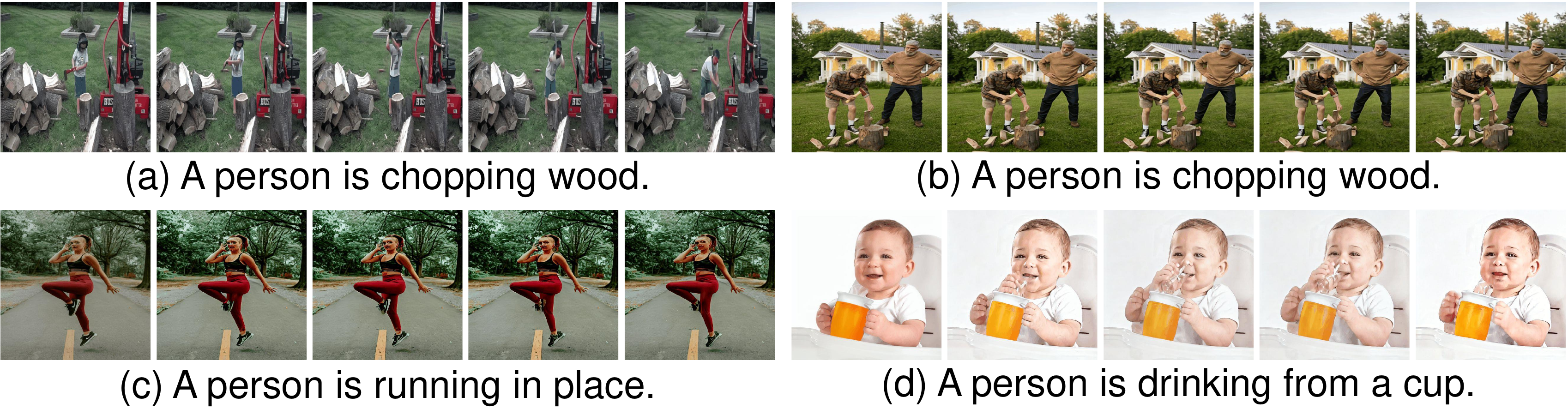}
\caption{Failure cases of \sysname{}.}
\label{fig:limitations}
\end{figure*}

\subsection{Experiments on UCF Sports Action Dataset}
\label{appendix:ucf_sports}
To evaluate the effectiveness of \sysname{} on additional datasets, we conducted experiments on the UCF Sports Action Dataset \cite{soomro2015action}, focusing on two actions: \textsf{\small golf swing} and \textsf{\small lifting}. Due to the limited number of videos in this dataset, we use only 6 \textsf{\small golf swing} videos and 4 \textsf{\small lifting} videos for training. For each class, we use the first frames of two videos for testing.

Table \ref{tab:main-ucf-sports} compares the performance of \sysname{} with baseline methods. \sysname{} achieves superior results on CD-FVD, Cosine RGB, and Cosine Flow, highlighting its ability to generate realistic motions. While DynamiCrafter performs better on Image Alignment and Temporal Consistency, this is primarily because it fails to animate the reference images and instead repeats it across frames, which represents a failure in animation. This limitation is further reflected in the poor scores of DynamiCrafter on CD-FVD and Cosine Flow. For Text Alignment, DreamVideo and TI2V-Zero outperform \sysname{}, but their inability to generate smooth transitions from reference images is evident from their low Image Alignment scores. These observations, consistent with results on the HAA dataset, demonstrate the effectiveness of \sysname{} in scenarios with fewer training videos.

\subsection{Experiments on Non-human Motion Videos}
\label{appendix:non_human}
To assess the performance of \sysname{} on non-human motion videos, we conducted experiments using two categories of natural motion: \textsf{\small firework} and \textsf{\small raining}. The videos were sourced from \cite{wu2023lamp}. For each category, we selected two videos for testing and used the remaining videos for training.

Table \ref{tab:main-non-human} presents a comparison between \sysname{} and baseline methods. \sysname{} achieves superior performance in CD-FVD, Cosine RGB, and Cosine Flow while not showing a obvious decline in CLIP scores. Although DynamiCrafter performs better in Image Alignment and Temporal Consistency, it struggles with CD-FVD and Cosine Flow. Similarly, DreamVideo excels in Text Alignment but performs poorly in Cosine RGB and Cosine Flow. These results indicate that \sysname{} can also animate images into videos depicting natural scene motion.

\section{Limitations}
\label{appendix:limitation}
Although \sysname{} can animate diverse reference images, it encounters challenges in accurately generating interactions involving human and objects, particularly when multiple objects are present. For example, in Figure \ref{fig:limitations}(a), while a chopping action is depicted, the object being chopped is not the wood. Furthermore, if the initial action states in the reference images differ noticeably in motion patterns from those in the training videos, the model may struggle with animation. For example, in Figure \ref{fig:limitations}(b), the initial action status suggests a small-scale motion for chopping wood, which differs from the large-scale motion in training videos; in Figure \ref{fig:limitations}(c), the knee elevation motion contrasts with the steadier motion of running in place observed in the training videos; and in Figure \ref{fig:limitations}(d), a baby holding a cup with both hands deviates from the adult actions in the training videos, where one hand is used to hold the cup while drinking water. These results suggest that the model still lacks a thorough understanding of motion and interactions. Leveraging advanced multi-modal large language models to improve the understanding of human-object interactions could be a promising approach to addressing these challenges.

\section{Ethics Statement}
\label{appendix:ethics}
We firmly oppose the misuse of generative AI for creating harmful content or spreading false information. We do not assume any responsibility for potential misuse by users. Nonetheless, we recognize that our approach, which focuses on animation human images, carries the risk of potential misuse. To address these risks, we are committed to maintaining the highest ethical standards in our research by complying with legal requirements and protecting privacy.



\end{document}